\documentclass{article}

\usepackage[accepted]{mlsys2025}

\usepackage[]{hyperref}
\usepackage{xspace}

\usepackage{multirow}
\usepackage{makecell}
\usepackage[group-separator={,}]{siunitx}
\usepackage{xfrac}

\usepackage{amsmath,amsfonts,bm}

\def\eqref#1{equation~\ref{#1}}

\def\1{\bm{1}}

\def\ry{{\textnormal{y}}}

\def\rvz{{\mathbf{z}}}

\def\va{{\bm{a}}}
\def\vb{{\bm{b}}}

\def\vw{{\bm{w}}}
\def\vx{{\bm{x}}}

\def\vz{{\bm{z}}}

\def\mW{{\bm{W}}}

\DeclareMathAlphabet{\mathsfit}{\encodingdefault}{\sfdefault}{m}{sl}
\SetMathAlphabet{\mathsfit}{bold}{\encodingdefault}{\sfdefault}{bx}{n}

\def\sD{{\mathbb{D}}}

\def\sK{{\mathbb{K}}}

\def\sM{{\mathbb{M}}}

\def\sS{{\mathbb{S}}}

\usepackage{amsmath}
\usepackage{mathtools}
\usepackage{amsthm}
\usepackage[ruled,noend]{algorithm2e}
\theoremstyle{plain}

\newtheorem{proposition}{Proposition}
\newtheorem{lemma}{Lemma}

\theoremstyle{definition}
\newtheorem{definition}{Definition}

\theoremstyle{remark}
\newtheorem{remark}{Remark}

\usepackage[capitalize,noabbrev]{cleveref}

\newcommand{\design}{FedProphet\xspace} 
\newcommand{\server}{[\textbf{Server}]:\xspace}
\newcommand{\client}{[\textbf{Client}]:\xspace}

\usepackage{pifont}
\usepackage{paralist}

\mlsystitlerunning{\design: Memory-Efficient Federated Adversarial Training via Robust and Consistent Cascade Learning}

\begin{document}

\twocolumn[
\mlsystitle{\design: Memory-Efficient Federated Adversarial Training via Robust and Consistent Cascade Learning}



\mlsyssetsymbol{equal}{*}

\begin{mlsysauthorlist}
\mlsysauthor{Minxue Tang}{equal,duke}
\mlsysauthor{Yitu Wang}{equal,duke}
\mlsysauthor{Jingyang Zhang}{duke}
\mlsysauthor{Louis DiValentin}{ac}
\mlsysauthor{Aolin Ding}{ac}
\mlsysauthor{Amin Hass}{ac}
\mlsysauthor{Yiran Chen}{duke}
\mlsysauthor{Hai ``Helen'' Li}{duke}
\end{mlsysauthorlist}

\mlsysaffiliation{duke}{Duke University}
\mlsysaffiliation{ac}{Accenture Cyber Labs}

\mlsyscorrespondingauthor{Yiran Chen}{yiran.chen@duke.edu}
\mlsyscorrespondingauthor{Hai ``Helen'' Li}{hai.li@duke.edu}

\mlsyskeywords{Federated Learning, Adversarial Training, Cascade Learning, Training Efficiency}

\vskip 0.3in



\begin{abstract}
Federated Adversarial Training (FAT) can supplement robustness against adversarial examples to Federated Learning (FL), promoting a meaningful step toward trustworthy AI. However, FAT requires large models to preserve high accuracy while achieving strong robustness, incurring high memory-swapping latency when training on memory-constrained edge devices. Existing memory-efficient FL methods suffer from poor accuracy and weak robustness due to inconsistent local and global models. In this paper, we propose \design, a novel FAT framework that can achieve memory efficiency, robustness, and consistency simultaneously. \design reduces the memory requirement in local training while guaranteeing adversarial robustness by adversarial cascade learning with strong convexity regularization, and we show that the strong robustness also implies low inconsistency in \design. We also develop a training coordinator on the server of FL, with Adaptive Perturbation Adjustment for utility-robustness balance and Differentiated Module Assignment for objective inconsistency mitigation. \design significantly outperforms other baselines under different experimental settings, maintaining the accuracy and robustness of end-to-end FAT with 80\% memory reduction and up to 10.8$\times$ speedup in training time.
\end{abstract}
]

\printAffiliationsAndNotice{\mlsysEqualContribution} 


\section{Introduction}
With the rapid development of data-gluttonous artificial intelligence (AI), concerns about training data privacy also arise. Federated Learning (FL) is proposed as a distributed machine learning paradigm to provide a strong privacy guarantee \cite{konevcny2015federated,konevcny2016federated}. FL pushes model training to local edge devices (denoted as clients in FL) and only aggregates locally trained models, which avoids privacy leakage in data gathering and transmission.

While FL can offer privacy benefits, it cannot guarantee robustness against adversarial examples that are deemed as another threat to AI systems. Previous studies have shown that AI models are usually sensitive to small input perturbations, and a manipulated imperceptible noise can cause catastrophic errors in the outputs \cite{goodfellow2014explaining}. To achieve adversarial robustness, adversarial training is proposed to train the model with adversarially perturbed training data \cite{madry2017towards}, and federated adversarial training (FAT) is also proposed to complement FL with adversarial training \cite{zizzo2020fat}.

However, adversarial training will sacrifice the model performance as a trade-off between the utility (accuracy) and the robustness \cite{wang2021convergence}. Therefore, a larger model with higher capacity is required to achieve high accuracy and strong robustness simultaneously. Since most clients in cross-device FL are resource-constrained edge devices such as IOT devices and mobile phones \cite{kairouz2019advances}, not all the clients have sufficient memory to train a large model demanded by FAT.

We are still able to train a model that exceeds the memory capacity by swapping the model parameters and intermediate features between the internal memory (e.g., CPU RAM or GPU memory) and the external storage (e.g., SSD)~\cite{rajbhandari2020zero,wang2023zero++}. When the internal memory is significantly smaller than the memory requirement for training the whole model, frequent memory swapping during each forward and backward propagation can incur high latency. The latency introduced by memory swapping becomes more significant in adversarial training, where additional backward and forward propagations are usually required to generate adversarially perturbed training data \cite{madry2017towards}.

Some previous studies explored memory-efficient federated learning frameworks that allow resource-constrained clients to train a smaller local model or a sub-model of the large global model, and aggregate the heterogeneous models with knowledge distillation \cite{lin2020ensemble,cho2022heterogeneous} or partial average \cite{diao2020heterofl,alam2022fedrolex}. However, previous methods cannot achieve high accuracy and strong robustness when being applied in FAT because of the \textbf{objective inconsistency}, i.e., clients train different local models from the global model. Specifically, the robustness of the local models does not sufficiently lead to the robustness of the global model with a different architecture. In addition, heterogeneous local models cause higher variance in the local model updates, which can lead to divergence of FAT since FAT is shown to be more unstable than standard FL \cite{zizzo2020fat,shah2021adversarial}.

In this paper, we propose \design, a memory-efficient FAT framework that can achieve strong adversarial robustness and low objective inconsistency simultaneously. 
Specifically, on the client side, we propose robust and consistent cascade learning which partitions a large global model into cascaded small modules such that the memory-constrained clients can train each module one by one without memory swapping. We incorporate adversarial training and strong convexity regularization into vanilla cascade learning \cite{belilovsky2020decoupled} to guarantee adversarial robustness, and our theoretical analysis further shows that strong adversarial robustness of the cascaded modules also implies low objective inconsistency in cascade learning. Thus, our adversarial cascade learning can also achieve low inconsistency. 
On the server side, we develop a training coordinator with two components: (a) Adaptive Perturbation Adjustment automatically controls the intermediate adversarial perturbation magnitude in adversarial cascade learning to attain a better utility-robustness balance and more stable convergence during training; (b) Differentiated Module Assignment further reduces the objective inconsistency by allowing ``prophet'' clients who have sufficient computational resources to train additional ``future'' modules together with the current module. In summary:


\begin{asparadesc}
    \item[(1)] \textbf{Framework:} We propose \design, a FAT framework with memory efficiency, adversarial robustness, and objective consistency.

    \item[(2)] \textbf{Client Side:} We develop adversarial cascade learning with strong convexity regularization to guarantee the robustness of the whole model. We theoretically demonstrate that the robustness achieved by our method also implies low objective inconsistency in cascade learning.

    \item[(3)] \textbf{Server Side:} We develop a training coordinator, with Adaptive Perturbation Adjustment that can attain better accuracy-robustness balance, and Differentiated Module Assignment that can further reduce the objective inconsistency without sacrificing efficiency.

    \item[(4)] \textbf{Experiments:} Under different settings, \design shows significantly higher accuracy and adversarial robustness than previous memory-efficient federated learning methods, maintaining almost the same utility and robustness as end-to-end FAT while saving $80\%$ memory and achieving up to $10.8\times$ speedup in training time\footnote{Our codes are available at \url{https://github.com/Yoruko-Tang/FedProphet.git}}.
\end{asparadesc}

\section{Related Works and Preliminaries}
\subsection{Federated Learning}
Federated Learning (FL) is a distributed learning framework, where different devices (i.e., clients) collaboratively train a model $\vw$ that can minimize the empirical task loss $L$ \cite{konevcny2015federated,mcmahan2017communication}:
\begin{equation}\label{eq:fl}
\begin{aligned}
\min_{\vw}&\quad L(\vw)=\sum_{k=1}^N q_kL_k(\vw), \\
\quad\text{where}&\quad L_k(\vw) = \frac{1}{\vert\sD_k\vert}\sum_{(\vx,y)\in \sD_k} l(\vx,y;\vw).
\end{aligned}
\end{equation}
$\sD_k$ with size $|\sD_k|=q_k\sum_i|\sD_i|$ is the local dataset of client $k$. The local dataset is never shared with others such that privacy is preserved in FL. FedAvg is the first and the most popular FL framework, with multi-step local SGD and periodical model average \cite{mcmahan2017communication}. 

One challenge in FL is the heterogeneous clients \cite{kairouz2019advances,li2020federated}, including statistical heterogeneity (non-I.I.D. and unbalanced local data) \cite{karimireddy2019scaffold,wang2020tackling,tang2022fedcor,zhang2023fed} and systematic heterogeneity (various computational resources) \cite{li2018federated,tian2022harmony,sun2022fedsea}. This paper mainly focuses on the systematic heterogeneity among clients, especially clients with insufficient memory.
Some recent studies propose \textit{Knowledge-distillation FL} where clients train heterogeneous models based on their computational resources, and the heterogeneous models are aggregated by knowledge distillation instead of average on the server \cite{lin2020ensemble,cho2022heterogeneous}. 
Some other studies develop \textit{Partial-training FL} where each client trains a sub-model extracted from the global large model, and the server aggregates the sub-models into the global large model by partial average \cite{caldas2018expanding,diao2020heterofl,alam2022fedrolex}. Though both knowledge-distillation FL and partial-training FL have been demonstrated effective in standard FL, they fail to tackle the objective inconsistency, leading to poor robustness and convergence in FAT.

\subsection{Adversarial Training}
It is well known that the performance of deep neural networks can be dramatically undermined by adversarial examples, which are generated by adding small perturbations to the clean data \cite{goodfellow2014explaining,yang2020dverge}. To improve the adversarial robustness, Adversarial Training (AT) is proposed. In contrast to standard training (ST) that simply minimizes the empirical task loss, AT solves a min-max problem during training: 
\begin{align}
    \min_{\vw}\max_{\boldsymbol{\delta}:\Vert \boldsymbol{\delta} \Vert\le \epsilon} l(\vx+\boldsymbol{\delta},y;\vw).\label{eq:at}
\end{align}
AT alternatively solves the inner maximization and the outer minimization such that the model becomes insensitive to the small perturbation $\boldsymbol{\delta}$ in the input $\vx$ \cite{croce2020reliable,wong2020fast}. For example, PGD-$n$ AT \cite{madry2017towards} conducts $n$ steps of projected gradient accent on $\boldsymbol{\delta}$ for the inner maximization, and then the perturbed inputs are used for one-step gradient descent on the model parameter $\vw$. 
AT with a larger $n$ usually confers stronger robustness to the model, but also incurs more forward and backward propagations \cite{wong2020fast}.

For the analysis purpose, we define $(\epsilon,c)$-robustness: 
\begin{definition}
A model $\vw$ is $(\epsilon,c)$-robust in a loss function $l$ at input $\vx$ if $\forall \boldsymbol{\delta}\in\{\boldsymbol{\delta}:\Vert \boldsymbol{\delta} \Vert \le \epsilon\}$,
\begin{align}
    l(\vx+\boldsymbol{\delta},y;\vw)-l(\vx,y;\vw) \le c.
\end{align}
\end{definition}

\subsection{Cascade Learning}
\begin{figure}[t]
\centering
\includegraphics[width=0.95\linewidth]{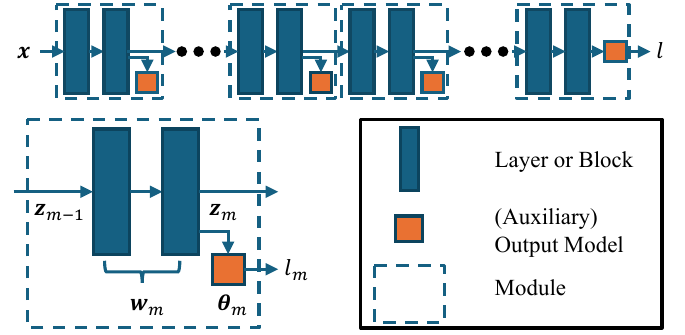}
\vspace{-0.5em}
\caption{An illustration of Cascade Learning. }
\label{fig:cascade}
\vspace{-1em}
\end{figure}

Cascade Learning (also known as Decoupled Learning) is proposed to reduce the memory requirement for training a large model \cite{hettinger2017forward,marquez2018deep,belilovsky2020decoupled}. As illustrated in \cref{fig:cascade}, Cascade Learning partitions a large neural network into cascaded small modules and trains the modules one by one in the forward order. Each module $\vw_m$ is trained  with an early exit loss $l_m$ provided by an auxiliary output model $\boldsymbol{\theta}_m$:
\begin{equation}
    \min_{\vw_m,\boldsymbol{\theta}_m} L_m(\vw_m,\boldsymbol{\theta}_m)
    = \mathbb E \left[l_m(\rvz_{m-1},\ry;\vw_m,\boldsymbol{\theta}_m)\right].
\end{equation}
After the current module $\vw_m$ converges, it is fixed as $\vw_m^*$, and the output features $\vz_m=\vz_m(\vz_{m-1};\vw_m^*)$ are used to train the next module $(\vw_{m+1},\boldsymbol{\theta}_{m+1})$.

However, vanilla cascade learning is shown to have inferior performance compared to end-to-end training because of the \textbf{objective inconsistency}, i.e., each module is independently trained with the early exit loss $l_m$ that is different from the joint loss $l$ of the whole model. Since $\nabla l_m\neq\nabla l$, each module converges to sub-optimum in the independent training \cite{wang2021revisiting}. In addition, the robustness of each module in the early exit loss does not sufficiently lead to the robustness of the whole model in the joint loss.

\section{Motivations}

\textbf{The resource-constrained edge devices are unable to afford federated adversarial training (FAT).} Due to the utility-robustness trade-off, FAT requires large models to achieve high accuracy and strong robustness simultaneously as shown in \cref{tab:at_size} \cite{wang2021convergence}. However, the common edge devices in cross-device FL scenarios are IOT devices, mobile phones, and laptops, which may not have sufficient memory capacity to afford training a large model (as shown in \cref{fig:memconsp}) \cite{kairouz2019advances}. 

\begin{table}[t]
\centering
\caption{Results of FAT in CIFAR-10 and Caltech-256 with different sizes of models. We use CNN3/VGG16 as the small model ($1\times$ memory)/large model ($5\times$ memory) in CIFAR-10, and CNN4/ResNet34 in Caltech-256. ``Large-PT'' adopts a partial-training FL method FedRolex \cite{alam2022fedrolex}.}
\label{tab:at_size}
\resizebox{\linewidth}{!}{
\begin{tabular}{c|cc|cc}
\hline
\multirow{2}{*}{Model (Mem)} & \multicolumn{2}{c|}{CIFAR-10}       & \multicolumn{2}{c}{Caltech-256}    \\ 
                       & Clean Acc. & Adv. Acc. & Clean Acc. & Adv. Acc. \\ \hline
Small ($1\times$)      & $66.57\%$  & $54.33\%$ & $25.64\%$  & $13.49\%$ \\
Large ($5\times$)      & $79.74\%$  & $56.76\%$ & $46.56\%$  & $19.76\%$\\ 
Large-PT ($1\times$) & $67.14\%$  & $54.13\%$ & $30.18\%$  & $11.78\%$\\ \hline
\end{tabular}
}
\vspace{-1em}
\end{table}

\begin{figure}[t]
\centering
\includegraphics[width=\linewidth]{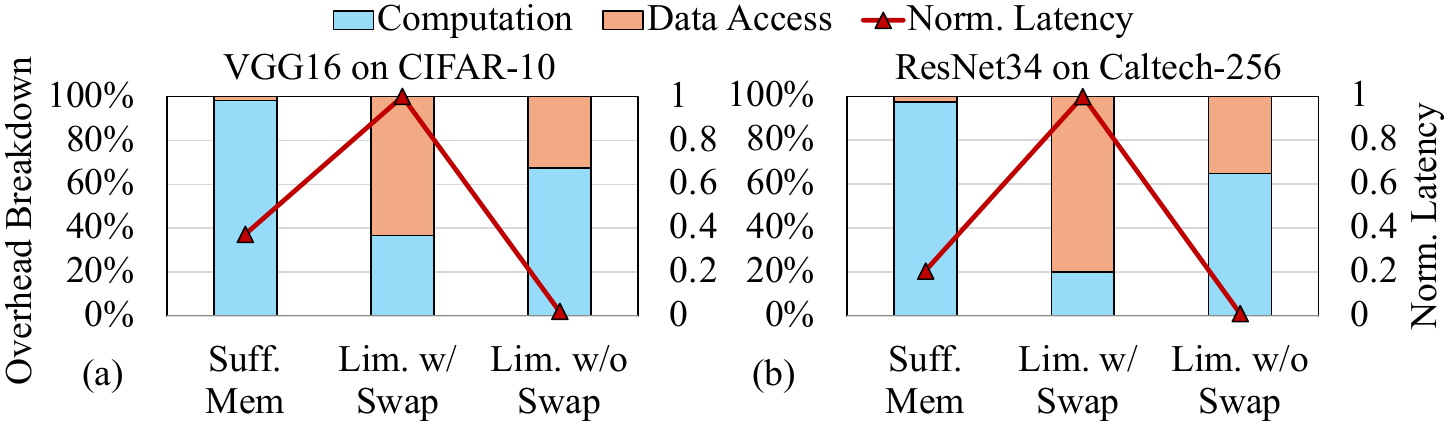}
\vspace{-2em}
\caption{The local training overhead breakdown and latency in two workloads, (a) VGG16 on CIFAR-10 and (b) ResNet34 on Caltech-256. ``Suff. Mem'' denotes training with sufficient memory resources and ``Lim. w/ Swap'' denotes training with 20\% memory and adopting memory swapping. ``Lim. w/o Swap'' trains with 20\% memory and FedRolex \cite{alam2022fedrolex}.}
\label{fig:mov_data_access}
\vspace{-1em}
\end{figure}

\begin{figure*}[t]
\centering
\includegraphics[width=0.95\linewidth]{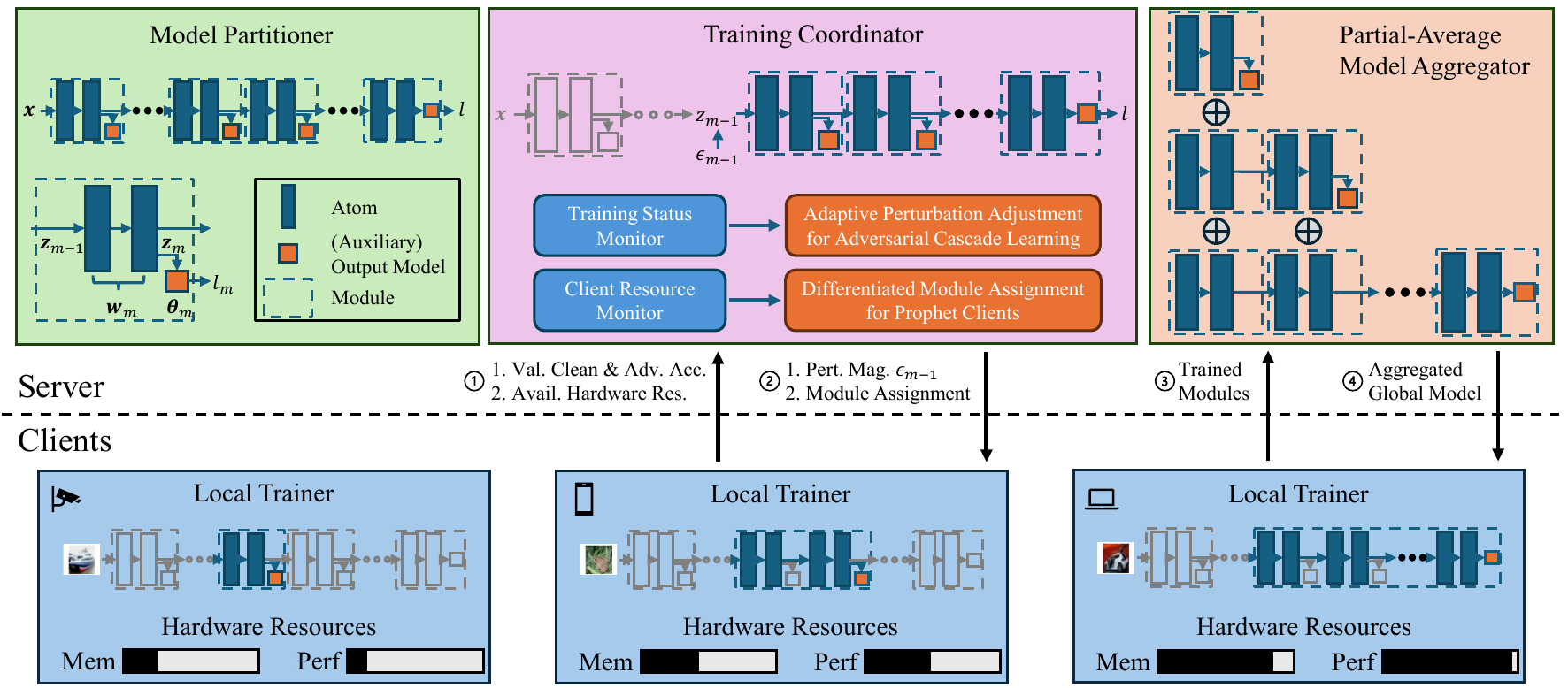}
\vspace{-0.5em}
\caption{A framework of \design. We formalize the framework in \cref{alg:fedprophet}. }
\label{fig:framework}
\vspace{-0.5em}
\end{figure*}

\textbf{Memory swapping introduces high data access overhead in FAT.} We can train a large model on a memory-constrained device by ``memory swapping'', which offloads/fetches the model parameters, optimizer states, and activations, to/from the external storage during training \cite{wang2023zero++}. However, memory swapping leads to high data access latency that can dominate the FAT workload as shown in \cref{fig:mov_data_access}. The high data access latency is usually caused by high software driver management overhead and low storage I/O bandwidth. Furthermore, FAT incurs more forward and backward propagations to solve the min-max problem in \cref{eq:at}, which increases the frequency of memory swapping.

\textbf{Previous memory-efficient FL methods fail to maintain high accuracy and robustness in FAT.} Prior memory-efficient FL methods can avoid memory swapping when training a large model \cite{lin2020ensemble,diao2020heterofl,cho2022heterogeneous,alam2022fedrolex}, but they show low clean and adversarial accuracy (exemplified by ``Large-PT'' in \cref{tab:at_size}). This is attributed to \textbf{objective inconsistency}, namely, memory-constrained clients locally train different small models from the large global model, leading to suboptimal local model updates and a gap between the local model robustness and the global model robustness.


\section{Overview}

Based on the motivations above, we propose \design, a memory-efficient FAT framework that can avoid memory swapping while maintaining utility and robustness when training a large model. As shown in \cref{fig:framework}, \design consists of client-side local trainers (\cref{sec:client}) and server-side model partitioner, training coordinator, and model aggregator (\cref{sec:server}). At the beginning of the FL process, the model partitioner partitions a predefined large global model into small modules that satisfy the memory constraints on clients. After model partitioning, we have four steps in each communication round of FL: \ding{172} Clients upload the validation (clean and adversarial) accuracy of and their available hardware resources (memory and performance) to the server; \ding{173} The training coordinator on the server adjusts the perturbation magnitude $\epsilon_{m-1}$ based on the validation accuracy (Adaptive Perturbation Adjustment), and determines which modules should be trained by each client based on their available hardware resources (Differentiated Module Assignment); \ding{174} Each client conducts adversarial training on the assigned modules with the strong-convexity regularized loss, and uploads the trained modules to the server; \ding{175} The server conducts partial average to aggregate the trained modules, and broadcasts the aggregated global model back to the clients.


\section{Local Client Design}\label{sec:client}

In this section, we introduce the local trainer on clients. 
We first introduce how a client adversarially trains a module with strong convexity regularization to ensure the robustness of the backbone model in \cref{subsec:amt}. In \cref{subsec:lowcons}, we will uncover the relationship between objective inconsistency and robustness, which demonstrates that the adversarial cascade learning we propose can also achieve low objective inconsistency in addition to strong robustness.

\subsection{Adversarial Cascade Learning with Strong Convexity Relularization}\label{subsec:amt}

\begin{figure}[t]
\centering
\includegraphics[width=0.95\linewidth]{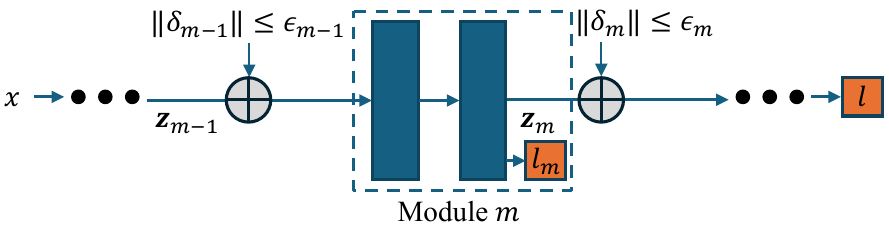}
\vspace{-0.5em}
\caption{An illustration of Adversarial Cascade Learning. }
\label{fig:amt}
\vspace{-1em}
\end{figure}

\paragraph{Sufficient Condition for Backbone Robustness.} A client can conduct adversarial training on module $m$ by adding adversarial perturbation to its input $\vz_{m-1}$, as shown in \cref{fig:amt}. However, since the module is trained with the early exit loss $l_m$ that is not equivalent to the joint loss $l$ of the backbone model, the robustness of the module in the early exit loss does not sufficiently lead to the robustness of the backbone model in the joint loss. The following proposition gives a sufficient condition for the robustness of the backbone model that consists of $M$ cascaded modules:
\begin{proposition}\label{prop:robustness}
The backbone model $(\vw_1\circ\cdots\circ \vw_M)$\footnote{We use $\va\circ\vb$ to denote a cascade of two modules $\va$ and $\vb$, inputting in $\va$ and outputting from $\vb$.} have $(\epsilon_0,c_M)$-robustness in the joint loss $l$, if for every module $m<M$, we have a finite upper bound $\epsilon_m$ such that
\begin{align}
    &\max_{\Vert\boldsymbol{\delta}_{m-1}\Vert\le \epsilon_{m-1}} \Vert \Delta\vz_{m}\Vert  \le \epsilon_m,\\
    \text{where}\quad &\Delta\vz_{m} = \vz_m(\vz_{m-1}+\boldsymbol{\delta}_{m-1})-\vz_m(\vz_{m-1}),
\end{align}
and the last module has $(\epsilon_{M-1},c_M)$-robustness in $l_M=l$.
\end{proposition}
\cref{prop:robustness} can be easily proved by induction on the number of modules. It shows that the backbone robustness can be achieved by the module robustness, with an upper bound of the perturbation on the output feature of each module. A straightforward method to ensure this upper bound is conducting adversarial training with the perturbation magnitude as the loss function, namely,
\begin{align}
    \min_{\vw_m}\max_{\Vert\boldsymbol{\delta}_{m-1}\Vert\le \epsilon_{m-1}}\Vert \Delta \vz_m\Vert.
\end{align}
However, calculating the gradient $\nabla_{\vw_m}\Vert \Delta \vz_m\Vert$ by backward propagation requires memory to store additional intermediate results of $\vz_{m-1}+\boldsymbol{\delta}_{m-1}$, which equivalently doubles the batch size during training. Therefore, this method is infeasible in the memory-constrained scenarios.

\paragraph{Strong Convexity Regularization.} The following lemma (proved in \cref{proof:lemma_robustness}) provides an alternative method to upper bound the perturbation on $\vz_m$, by making the early exit loss function strongly convex on $\vz_m$ and conducting adversarial training with it:
\begin{lemma}\label{lemma:robustness}
    If $l_m$ is $\mu_m$-strongly convex on $\vz_m$ and module $m$ is $(\epsilon_{m-1},c_m)$-robust in $l_m$, we have
    \begin{equation}
        \begin{aligned}
        &\max_{\Vert\boldsymbol{\delta}_{m-1}\Vert\le \epsilon_{m-1}}\Vert \Delta \vz_m\Vert_2\\ 
        \le &\frac{\Vert \nabla_{\vz_m} l_m(\vz_m,y)\Vert_2}{\mu_m}+\sqrt{\frac{2c_m}{\mu_m}+\frac{\Vert \nabla_{\vz_m} l_m(\vz_m,y)\Vert_2^2}{\mu_m^2}}.
        \end{aligned}
    \end{equation}
\end{lemma}

According to \cref{prop:robustness} and \cref{lemma:robustness}, we propose the following two designs for each module $m$ to attain robustness of the backbone model in the joint loss $l$:
\begin{asparadesc}
    \item[(1)] Use a linear layer $\boldsymbol{\theta}_m=\{\mW_m,\vb_m\}$ (i.e., a fully connected layer) as the auxiliary model.
    \item[(2)] Conduct adversarial training on the early exit loss with strong convexity regularization:
    \begin{equation}\label{eq:module_loss}
    \begin{aligned}
        \min_{\vw_m,\boldsymbol{\theta}_m}&\max_{\Vert\boldsymbol{\delta}_{m-1}\Vert\le \epsilon_{m-1}} l_m(\vz_{m-1}+\boldsymbol{\delta}_{m-1},y;\vw_m,\boldsymbol{\theta}_m)\\
        =\min_{\vw_m,\boldsymbol{\theta}_m}&\max_{\Vert\boldsymbol{\delta}_{m-1}\Vert\le \epsilon_{m-1}}
        \bigg[\frac{\mu}{2}\Vert \vz_m(\vz_{m-1}+\boldsymbol{\delta}_{m-1};\vw_m)\Vert_2^2\\
        &+l_{\text{CE}}(\mW_m^T \vz_m(\vz_{m-1}+\boldsymbol{\delta}_{m-1};\vw_m)+\vb_m,y)\bigg].
    \end{aligned}
    \end{equation}
\end{asparadesc}
The auxiliary model with a single fully connected layer and the cross-entropy (CE) loss can guarantee convexity but not strong convexity. Thus, we add an $\ell_2$ regularizer on $\vz_m$ to attain strong convexity. By adopting this strong-convexity regularized loss in all the modules, we can achieve upper bounded perturbation on the output feature of each module, thereby ensuring the sufficient condition for backbone robustness in \cref{prop:robustness}.

\begin{remark}
Notice that the weak convexity of the linear layer is not sufficient to upper bound the perturbation on $\vz_m$. The perturbation on $\vz_m$ that lies in the null space of $\mW_m$ can be arbitrarily large while not changing the output.
\end{remark}

\subsection{Robustness-Consistency Relationship in Adversarial Cascade Learning}\label{subsec:lowcons}
While we have guaranteed the backbone robustness with adversarial cascade learning, the poor performance caused by the objective inconsistency of cascade learning is not addressed. Since the early exit loss $l_m$ is not equivalent to the joint loss $l$, the gradient $\nabla_{\vw_m} l_m$ provided by the early exit loss is not the same as $\nabla_{\vw_m} l$ either, leading to suboptimal model updates. To mitigate the inconsistency, we need to reduce the gradient difference $\Vert \nabla_{\vw_m} l- \nabla_{\vw_m} l_m\Vert$.

There is usually no upper bound on the gradient difference since the auxiliary model can be arbitrarily different from the backbone model. However, under the adversarial cascade learning with both module robustness and backbone model robustness, we can derive an upper bound on the gradient difference with the following lemma:
\begin{lemma}\label{lemma:consist}
    If the early exit loss $l_m$ has $\beta_m$-smoothness and $(\epsilon_m,c_m)$-robustness on $\vz_m$, the joint loss $l$ has $\beta_m'$-smoothness and $(\epsilon_m,c_M)$-robustness on $\vz_m$, we have
    \begin{equation}
    \begin{aligned}
        &\Vert\nabla_{\vw_m} l-\nabla_{\vw_m} l_m\Vert_2\\
        \le&\left\Vert \frac{\partial \vz_m}{\partial \vw_m}\right\Vert_2\sqrt{2(c_m+c_M)(\beta_m+\beta_m')}.
    \end{aligned}
    \end{equation}
\end{lemma}
\cref{lemma:consist} is proved in \cref{proof:lemma_consist}. An intuitive explanation is that if both $l_m$ and $l$ are not sensitive to the perturbation on $\vz_m$ (i.e., smooth and robust), their gradients on $\vz_m$ are close to $0$ simultaneously and thus the gradient difference is small. A previous study has shown that the robustness of a deep neural network also implies smoothness \cite{moosavi2019robustness}. Thus, we attain small $\beta_m$ and $c_m$ with the module robustness of module $m$, and we attain small $\beta_m',c_M$ with the backbone robustness of the whole model. In conclusion, our adversarial cascaded learning can also mitigate objective inconsistency.



\section{Central Server Design}\label{sec:server}

In this section, we introduce how the server coordinates the training in \design with three components: model partitioner (\cref{subsec:mp}), training coordinator (Adaptive Perturbation Adjustment in \cref{subsec:aapa} and Differentiated Module Assignment in \cref{subsec:dpma}), and partial-average model aggregator (\cref{subsec:pama}). 

\subsection{Memory-constrained Model Partition}\label{subsec:mp}
The model partitioner partitions the backbone model into cascaded modules and each module can be independently trained with an auxiliary model. To formalize the model partitioning, we first define the \textbf{``atom''} which cannot be further partitioned. An ``atom'' of a model is a layer or a block such that the backbone model is constructed as a plain cascade of multiple ``atoms'' ($a_1\circ\cdots\circ a_L$). For example, the ``atom'' of a plain neural network (e.g., VGG \cite{simonyan2014very}) is a single layer, while the ``atom'' of ResNet \cite{he2016deep} is a residual block. A module consists of several connected ``atoms'', with an extra auxiliary model appended to the end.

\begin{algorithm}[t]
\caption{Memory-constrained Model Partition}
\label{alg:mp}
\SetKwInput{KwRequire}{Require} 
\SetKwComment{Comment}{/* }{ */}
\KwRequire{The ``atom'' sequence ($a_1\circ\cdots\circ a_L$); Minimal reserved memory $R_{\text{min}}$}
Initialize $\sM=\emptyset,m=\emptyset$\;
\For{$i\le L$}{
\eIf{$\text{MemReq}(m\cup\{a_i\})<R_{\text{min}}$}{
Append $a_i$ to $m$\;
}{
Append $m$ to $\sM$\;
$m\leftarrow \{a_i\}$\;
}
}
Append $m$ to $\sM$\;
\KwResult{Model partition $\sM$}
\end{algorithm}



The key of the model partitioner is to ensure that the memory requirement for training each module does not exceed the minimal reserved memory $R_{\text{min}}$ among all the clients, such that the clients can train at least one module without memory swapping in any communication round. We adopt a greedy model partitioning method given in \cref{alg:mp}, which can achieve the least number of modules under the memory constraint. Specifically, we traverse each ``atom'' in the model and append it into one module until it reaches the memory constraint. The $\text{MemReq}(m)$ in \cref{alg:mp} is a function that returns the memory requirement for training the module $m$. In this paper, we adopt the methodology proposed by \citet{rajbhandari2020zero} to estimate the memory requirement, considering model parameters, gradients, optimizer states, and intermediate activations.

\subsection{Adaptive Perturbation Adjustment}\label{subsec:aapa}
When conducting adversarial training on module $m$, though \cref{prop:robustness} requires setting the perturbation constraint $\epsilon_{m-1}$ as the upper bound of $\Delta\vz_{m-1}$ to sufficiently guarantee the backbone robustness, we find that it is not necessary to use such a large perturbation magnitude in practice. On the one hand, a too-large perturbation magnitude on the intermediate features can cause a significant accuracy drop and even lead to divergence. On the other hand, a too-small perturbation magnitude cannot confer strong robustness to the backbone model. Therefore, it is essential to find an appropriate perturbation magnitude for each module to achieve the best utility-robustness trade-off.

Since the optimal perturbation constraint may differ in different modules, we propose \textbf{Adaptive Perturbation Adjustment} mechanism to automatically adjust the perturbation magnitudes during training. When module $m-1$ is fixed after convergence, we collect the largest perturbation $\Delta\vz_{m-1}$ on its output with the adversarial perturbation on its input $\vz_{m-2}$ from all clients. Then we set the constraint $\epsilon_{m-1}^{(t)}$ for training the next module $m$ based on the averaged perturbation magnitude on $\vz_{m-1}$ as follows\footnote{Notice that we do not adjust $\epsilon_0$ for the original data and always set it as a predefined value, e.g., \sfrac{8}{255}.}:
\begin{equation}\label{eq:eps}
\begin{aligned}
    \epsilon_{m-1}^{(t)} = \alpha_{m-1}^{(t)}\mathbb{E}\left[\max_{\Vert\boldsymbol{\delta}_{m-2}\Vert\le \epsilon_{m-2}^*}\Vert \Delta \rvz_{m-1}\Vert\right].
\end{aligned}
\end{equation}
Adaptive Perturbation Adjustment adaptively tunes the scaling factor $\alpha_{m-1}^{(t)}$ at each communication round $t$ to balance the utility and robustness. The foundation of this mechanism is that the ratio between the clean accuracy (accuracy on clean examples) and the adversarial accuracy (accuracy on adversarial examples) reveals the balance between utility and robustness, and this ratio should not change significantly when cascading one more module. Therefore, we monitor the ratio between the clean accuracy and the adversarial accuracy during training, and we adjust $\alpha_{m-1}^{(t)}$ by comparing the accuracy ratio of this module and the previous module as follows:
\begin{equation}\label{eq:ada_alpha}
    \alpha_{m-1}^{(t)}=
    \begin{cases}
        \alpha_{m-1}^{(t-1)} + \Delta\alpha,\quad\text{if }\frac{C_{m}^{(t)}}{A_{m}^{(t)}}>(1+\gamma)\frac{C_{m-1}^*}{A_{m-1}^*};\\
        \alpha_{m-1}^{(t-1)} - \Delta\alpha,\quad\text{if }\frac{C_{m}^{(t)}}{A_{m}^{(t)}}<(1-\gamma)\frac{C_{m-1}^*}{A_{m-1}^*};\\
        \alpha_{m-1}^{(t-1)},\quad\quad\quad\ \text{elsewhere}.
    \end{cases}
\end{equation}
$C_{m}^{(t)}$ and $A_{m}^{(t)}$ are the validation clean accuracy and adversarial accuracy of the cascaded modules $(\vw_1^*\circ\vw_2^*\circ\cdots\circ \vw_m^{(t)})$ at communication round $t$. $C_{m-1}^*$ and $A_{m-1}^*$ denote the final clean and adversarial accuracy of $(\vw_1^*\circ\vw_2^*\circ\cdots\circ \vw_{m-1}^*)$ when completing training module $m-1$ and fixing it. $\gamma$ is a small threshold constant, e.g., $0.05$ in our experiments. When the accuracy ratio is too large, which means that the clean accuracy is too high and the adversarial accuracy is too low, we increase the scaling factor $\alpha_{m-1}^{(t)}$ by a small constant $\Delta \alpha$ (e.g., $0.1$ in our experiments) to enhance the robustness, and vice versa. 


\subsection{Differentiated Module Assignment}\label{subsec:dpma}
Since different clients in federated learning may have different available hardware resources, it is possible that some of them can train multiple modules or even the whole backbone model at a time with sufficient memory and performance. Cascading more modules and training them together can equivalently reduce the number of modules and mitigate objective inconsistency in cascade learning \cite{wang2021revisiting}. 
Thus, we propose \textbf{Differentiated Module Assignment} in the training coordinator to fully utilize the computational resources of resource-sufficient clients.

Differentiated Module Assignment mechanism assigns different numbers of modules to clients in each communication round according to their real-time available resources. A client $k$ who is assigned multiple modules $\{m,m+1, \cdots M_k^{(t)}\}$ in round $t$ trains the cascaded modules jointly with the following loss:
\begin{equation}\label{eq:prophet_loss}
\begin{aligned}
    &l_k^{(t)}\left(\vz_{m-1},y;\vw_m,\cdots,\vw_{M_k^{(t)}},\boldsymbol{\theta}_{M_k^{(t)}}\right)\\ 
    = 
    &l_{\text{CE}}\left(\mW_{M_k^{(t)}}^T\vz_{M_k^{(t)}}(\vz_{m-1};\vw_m,\cdots,\vw_{M_k^{(t)}})+\vb_{M_k^{(t)}},y\right)\\
    &+\frac{\mu}{2}\Vert\vz_{M_k^{(t)}}(\vz_{m-1};\vw_m,\cdots,\vw_{M_k^{(t)}})\Vert_2^2,
\end{aligned}
\end{equation}
where $\vz_{M_k^{(t)}}(\vz_{m-1})$ calculates the feature $\vz_{M_k^{(t)}}$ by forward propagating the input $\vz_{m-1}$ through the cascaded modules ($\vw_m\circ \vw_{m+1}\circ \cdots\circ \vw_{M_k^{(t)}}$). Then the early exit loss provided by the auxiliary model $\boldsymbol{\theta}_{M_k^{(t)}}$ of the last assigned module $M_k^{(t)}$, together with the strong convexity regularization, is used to train the cascaded modules.

\begin{figure}[t]
\centering
\includegraphics[width=0.9\linewidth]{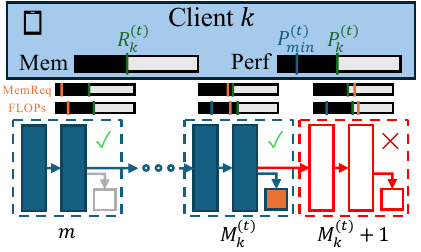}
\vspace{-0.5em}
\caption{An illustration of resource-constrained module assignment with memory and FLOPs constraints. }
\label{fig:assignment}
\vspace{-1em}
\end{figure}
\paragraph{Resource-constrained Module Assignment.} 
We now discuss how to assign modules to maximize the utilization of the resources of each client. When determining the module assignment for a client $k$, we choose the largest $M_k^{(t)}$ that satisfies both of the following two constraints:
\begin{asparadesc}
    \item[(1)] The total memory requirement of the assigned modules should not exceed the available memory $R_k^{(t)}$:
    \begin{equation}\label{eq:mem_cons}
        \text{MemReq}(\vw_m\circ \cdots\circ \vw_{M_k^{(t)}}\circ\boldsymbol{\theta}_{M_k^{(t)}})\le R_k^{(t)}.
    \end{equation}
    \item[(2)] Training the assigned modules on client $k$ should not take longer than training a single module $m$ on the slowest client:
    \begin{equation}\label{eq:flop_cons}
        \text{FLOPs}(\vw_m\circ \cdots\circ \vw_{M_k^{(t)}}\circ \boldsymbol{\theta}_{M_k^{(t)}})\le \frac{P_k^{(t)}}{P_{\text{min}}^{(t)}}\text{FLOPs}(\vw_m).
    \end{equation}
\end{asparadesc}
$\text{MemReq}()$ returns the memory requirement of the given model, and $\text{FLOPs}()$ returns the FLOPs for training the model. $P_k^{(t)}$ is the available performance of client $k$ at round $t$ and $P_{\text{min}}^{(t)}$ is the lowest available performance among clients who participate in the training of round $t$. As shown in \cref{fig:assignment}, we increase the number of modules that are assigned to a ``prophet'' client when both the memory constraint (\cref{eq:mem_cons}) and the FLOPs constraint (\cref{eq:flop_cons}) can be satisfied. The memory constraint avoids the data access latency incurred by memory swapping, while the FLOPs constraint minimizes the synchronization time of each communication round in FL by setting a hard limit on the training time of the assigned modules, which should be no more than the time of training only one module $m$ on the slowest client.

\begin{algorithm}[t]
\caption{\design}
\label{alg:fedprophet}
\SetKwInput{KwRequire}{Require} 
\SetKwComment{Comment}{/* }{ */}
\KwRequire{The initial model $\vw^{(0)}$; Minimal reserved memory $R_{\text{min}}$; Strong convexity hyperparameter $\mu$.}
\server partitions the model into $M$ modules $\vw = \{\vw_1,\cdots,\vw_M\}$ according to $R_{\text{min}}$ (Sec. \ref{subsec:mp})\;
\server Round $t\leftarrow 0$, broadcasts $\vw^{(0)}$\;
\For{Module $m=1,\cdots, M$}{
    \While{$\vw_m$ does not converge}{
        \If{$m>1$}{
            \server adjusts $\epsilon_{m-1}^{(t)}$ (Sec. \ref{subsec:aapa})\;
        }
        \For{each Client $k$}{
            \server assigns $M_k^{(t)}$ (Sec. \ref{subsec:dpma})\; 
            \client conducts adversarial training on $\{m,\cdots, M_k^{(t)}\}$ with $\epsilon_{m-1}=\epsilon_{m-1}^{(t)}$ and $l_m=l_k^{(t)}$ in Eq. (\ref{eq:module_loss}) (Sec. \ref{subsec:amt}\&\ref{subsec:dpma})\;
            \client uploads trained modules\;
        }
        \server aggregates modules (Sec. \ref{subsec:pama})\;
        \server Round $t\leftarrow t+1$, broadcasts $\vw^{(t)}$\;
    }
    [\textbf{All Clients}]: Fix $\vw_m^*\leftarrow\vw_m^{(t)},\epsilon_{m-1}^*\leftarrow \epsilon_{m-1}^{(t)}$\;
    \server Collects $\max_{\boldsymbol{\delta}_{m-1}} \Vert \Delta \vz_m\Vert$ from clients\;
}
\KwResult{Trained model $\vw^* = \{\vw_1^*,\cdots,\vw_M^*\}$.}
\end{algorithm}

\subsection{Partial-Average Model Aggregator}\label{subsec:pama}
With the Differentiated Module Assignment mechanism, the server needs to aggregate the updated local models with different numbers of modules from different clients. Similar to previous partial-training FL algorithms \cite{caldas2018expanding,diao2020heterofl,alam2022fedrolex}, we adopt partial average to aggregate local models with different numbers of modules. For each module  $m\le n\le M$, the module parameter $\vw_n$ is aggregated as:
\begin{equation}\label{eq:aggre_backbone}
    \vw_n^{(t+1)}=\frac{\sum_{k\in\sS_n^{(t)}}q_k\vw_{n,k}^{(t,E)}}{\sum_{k\in\sS_n^{(t)}}q_k}, \quad 
    \sS_n^{(t)}=\{k:M_k^{(t)}\ge n\},
\end{equation}
where $\sS_n^{(t)}$ is the set of clients who trained module $n$ in communication round $t$, and $\vw_{n,k}^{(t,E)}$ is the local module parameters trained by client $k$ for $E$ local iterations in this round. We also aggregate the auxiliary model $\boldsymbol{\theta}_n$ similarly:
\begin{equation}\label{eq:aggre_aux}
\boldsymbol{\theta}_n^{(t+1)}=\frac{\sum_{k\in\sK_n^{(t)}}q_k\boldsymbol{\theta}_{n,k}^{(t,E)}}{\sum_{k\in\sK_n^{(t)}}q_k},\quad
\sK_n^{(t)} = \{k:M_k^{(t)}=n\}.
\end{equation}

We summarize \design in \cref{alg:fedprophet}.

\section{Empirical Evaluation}\label{sec:exp}

\subsection{Experiment Setup}
\paragraph{Datasets and Statistical Heterogeneity.}
We adopt two popular image classification datasets, CIFAR-10 with 10 classes of $3\times 32\times 32$ images \cite{krizhevsky2009learning} and Caltech-256 with 256 classes of $3\times 224\times 224$ images \cite{griffin2007caltech}, for empirical evaluation. For both datasets, we partition the whole training set onto $N=100$ clients and we randomly select $C=10$ clients to participate in training at each round. We adopt the same statistical heterogeneity as in previous FAT literature \cite{shah2021adversarial}: On each client, $80\%$ training data belongs to around $20\%$ classes (i.e., 2 classes in CIFAR-10 and 46 classes in Caltech-256), and $20\%$ data belongs to the other classes.

\paragraph{Devices and Systematic Heterogeneity.} We collect device pools consisting of common edge devices, with details in \cref{apx:dev}.
When sampling the devices from the pool, we emulate two levels of systematic heterogeneity: \textbf{(a) balanced:} We sample different devices with equal probability; \textbf{(b) unbalanced:} We give a higher sampling probability for devices with smaller memory and lower performance. 
\cref{fig:memconsp} shows the distribution of memory and performance in the real-time device samplings.

\paragraph{Models and Evaluation Metrics.}
We conduct PGD-10 adversarial training \cite{madry2017towards} with VGG16 \cite{simonyan2014very} on CIFAR-10, and ResNet34 \cite{he2016deep} on Caltech-256. We report the test accuracy on clean data (Clean Acc.), and we conduct PGD-20 attack (PGD Acc. or Adv. Acc.) and Auto Attack (AA Acc.) \cite{croce2020reliable} to evaluate the robustness of the model. Following previous adversarial training literature \cite{zhang2019interpreting}, the perturbations on both training data and test data are bounded by $\ell_\infty$ norm with $\epsilon_0=\sfrac{8}{255}$. We report the training time (including computation time and data access time) as the efficiency metric. 

\paragraph{Baselines.} We compare \design with joint federated adversarial learning (jFAT) \cite{zizzo2020fat}, knowledge-distillation federated adversarial training (FedDF-AT \cite{lin2020ensemble}, FedET-AT \cite{cho2022heterogeneous}), partial-training federated adversarial training (HeteroFL-AT \cite{diao2020heterofl}, FedDrop-AT \cite{wen2022federated}, FedRolex-AT \cite{alam2022fedrolex}), and Federated Robustness Propagation (FedRBN) \cite{hong2023federated}. We provide a detailed description of the baselines in \cref{apx:base}.

\subsection{Performance of \design}


\begin{figure}[t]
\centering
\includegraphics[width=\linewidth]{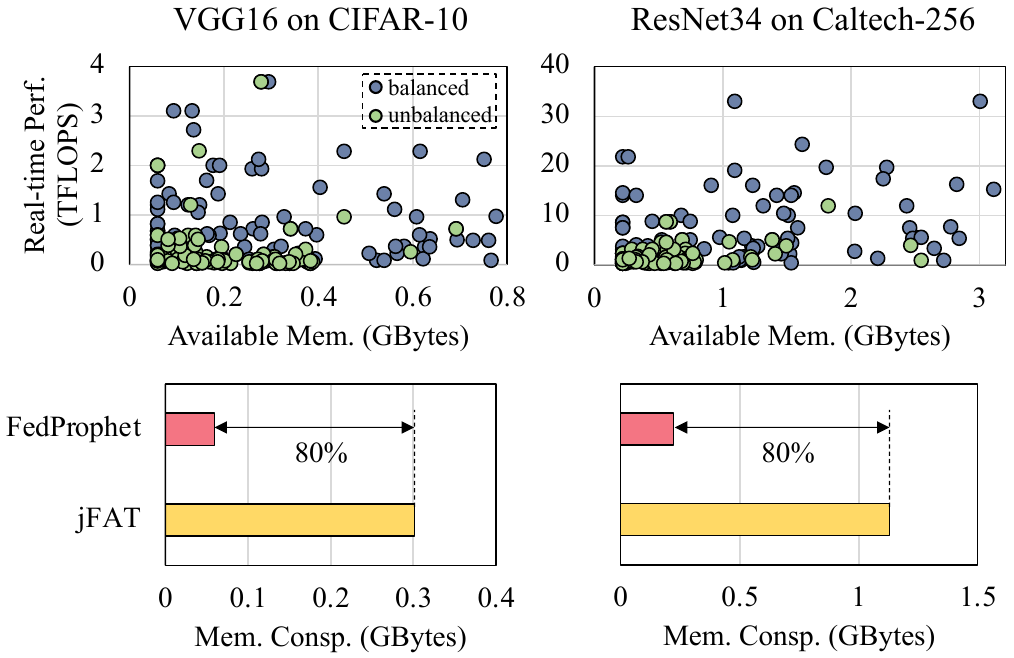}
\vspace{-2em}
\caption{The balanced (blue dots) /unbalanced (green dots) device samplings (upper), and the memory consumption of jFAT and FedProphet (lower) on two workloads.}
\label{fig:memconsp}
\vspace{-1em}
\end{figure}

\begin{table*}[t]
\centering
\caption{Clean Accuracy (Clean Acc.) and Adversarial Accuracy against PGD (PGD Acc.) and AutoAttack (AA Acc.). We highlight the best results among all methods besides jFAT which requires more memory or memory swapping when training.}
\label{table:main}
\resizebox{\linewidth}{!}{
\begin{tabular}{c|cccccc|cccccc}
\hline
Dataset      & \multicolumn{6}{c|}{CIFAR-10 ($32\times32$)}                                                                                                & \multicolumn{6}{c}{Caltech-256 ($224\times224$)}                                                                                            \\
Sys. Hetero. & \multicolumn{3}{c|}{balanced}                                               & \multicolumn{3}{c|}{unbalanced}                        & \multicolumn{3}{c|}{balanced}                                               & \multicolumn{3}{c}{unbalanced}                         \\ \hline
Method       & Clean Acc.       & PGD Acc.         & \multicolumn{1}{c|}{AA Acc.}          & Clean Acc.       & PGD Acc.         & AA Acc.          & Clean Acc.       & PGD Acc.         & \multicolumn{1}{c|}{AA Acc.}          & Clean Acc.       & PGD Acc.         & AA Acc.          \\ \hline
jFAT         & 79.74\%          & 56.76\%          & \multicolumn{1}{c|}{55.01\%}          & 79.74\%          & 56.76\%          & 55.01\%          & 46.56\%          & 19.76\%          & \multicolumn{1}{c|}{18.36\%}          & 46.56\%          & 19.76\%          & 18.36\%          \\ \hline
FedDF-AT     & 47.77\%          & 24.88\%          & \multicolumn{1}{c|}{18.72\%}          & 48.16\%          & 25.39\%          & 18.34\%          & 6.74\%           & 4.83\%           & \multicolumn{1}{c|}{4.10\%}           & 11.78\%          & 0.09\%           & 0\%              \\
FedET-AT     & 40.73\%          & 7.29\%           & \multicolumn{1}{c|}{5.12\%}           & 34.91\%          & 8.74\%           & 5.54\%           & 11.48\%          & 2.76\%           & \multicolumn{1}{c|}{2.44\%}           & 16.49\%          & 1.92\%           & 1.73\%           \\
HeteroFL-AT  & 51.63\%          & 39.36\%          & \multicolumn{1}{c|}{38.47\%}          & 55.25\%          & 43.05\%          & 41.96\%          & 27.80\%          & 8.70\%           & \multicolumn{1}{c|}{8.15\%}           & 9.43\%           & 3.04\%           & 2.87\%           \\
FedDrop-AT   & 65.92\%          & 54.21\%          & \multicolumn{1}{c|}{53.23\%}          & 63.26\%          & 53.21\%          & 52.61\%          & 27.10\%          & 11.87\%          & \multicolumn{1}{c|}{10.05\%}          & 11.68\%          & 6.54\%           & 5.20\%           \\
FedRolex-AT  & 67.14\%          & 54.13\%          & \multicolumn{1}{c|}{53.51\%}          & 66.44\%          & 53.25\%          & 52.00\%          & 30.18\%          & 11.78\%          & \multicolumn{1}{c|}{9.84\%}           & 12.51\%          & 5.80\%           & 4.81\%           \\
FedRBN       & \textbf{84.81\%} & 42.88\%          & \multicolumn{1}{c|}{39.82\%}          & \textbf{86.70\%} & 42.99\%          & 39.85\%          & \textbf{78.38\%} & 3.14\%           & \multicolumn{1}{c|}{0\%}           & \textbf{78.81\%} & 1.43\%           & 0\%              \\ \hline
\design       & 77.79\%          & \textbf{59.22\%} & \multicolumn{1}{c|}{\textbf{57.89\%}} & 76.47\%          & \textbf{59.51\%} & \textbf{58.64\%} & 47.07\%          & \textbf{19.10\%} & \multicolumn{1}{c|}{\textbf{18.11\%}} & 43.39\%          & \textbf{14.93\%} & \textbf{14.41\%} \\ \hline
\end{tabular}
}
\end{table*}

\begin{figure*}[t]
\centering
\includegraphics[width=\linewidth]{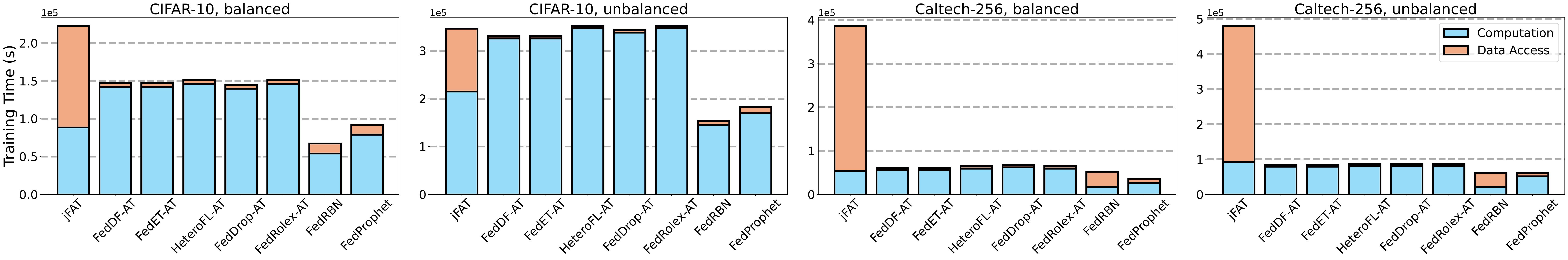}
\vspace{-2em}
\caption{Training time (including computation time and data access time) of baselines and \design. }
\label{fig:exp_time}
\vspace{-1em}
\end{figure*}

\paragraph{Memory Requirements.} We set the minimal reserved memory $R_{\text{min}}$ of memory-constrained clients to be around $20\%$ of the memory required for training the whole model, as shown by \cref{fig:memconsp}. Specifically, $R_{\text{min}}=60$ MBytes when training VGG16 (Requires $302\text{ MBytes}$) on CIFAR-10, and $R_{\text{min}}=224$ MBytes when training ResNet34 (Requires $1130\text{ MBytes}$) on Caltech-256. \design partitions both VGG16 and ResNet34 into $7$ modules with \cref{alg:mp} to avoid memory swapping. In other words, \design reduces the theoretical memory requirement by 80\% in comparison to jFAT. \cref{apx:part} provides more details of the model partition.

\paragraph{Utility and Robustness.} The utility (clean accuracy) and robustness (adversarial accuracy) of different methods are reported in \cref{table:main}. Compared to jFAT, only \design can maintain comparable or even higher utility and robustness simultaneously in all our settings. The objective inconsistency between the local models and the global model leads to suboptimal model updates and poor performance of previous memory-efficient baselines. Although FedRBN avoids objective inconsistency with homogeneous models and achieves high clean accuracy, it fails to attain strong robustness under high systematic heterogeneity where most clients cannot afford joint adversarial training \cite{hong2023federated}. \design guarantees backbone robustness while overcoming the objective inconsistency, attaining significantly better utility and robustness than these baselines.

\paragraph{Training Efficiency.} \cref{fig:exp_time} shows the total training time of different methods, including the computation time and data access time. The high frequency of memory swapping in jFAT when training the large backbone model incurs significant data access time and slows down the training procedure. The other memory-efficient methods avoid training the memory-exceeding large model and thus the data access time is much smaller than that of jFAT. However, since only a small part of the whole model is trained in each round, the memory-efficient methods usually require more communication rounds for convergence \cite{diao2020heterofl,wen2022federated,alam2022fedrolex}, as indicated by the higher computation time than jFAT. \design compensates for the extra communication round by adopting the FLOPs-constrained module assignment in \cref{eq:flop_cons}, which reduces the synchronization time in each communication round of FL. Thus, \design attains low data access time and low computation time simultaneously, with $2.4\times,1.9\times,10.8\times,7.7\times$ speedup in the total training time compared to jFAT in each setting respectively.

\subsection{Ablation Study of Components in \design}
We conduct ablation studies in this section to evaluate the functionality of each component in \design, including the client-side local trainer, the server-side model partitioner, and the server-side training coordinator.

\begin{figure}[t]
\centering
\includegraphics[width=\linewidth]{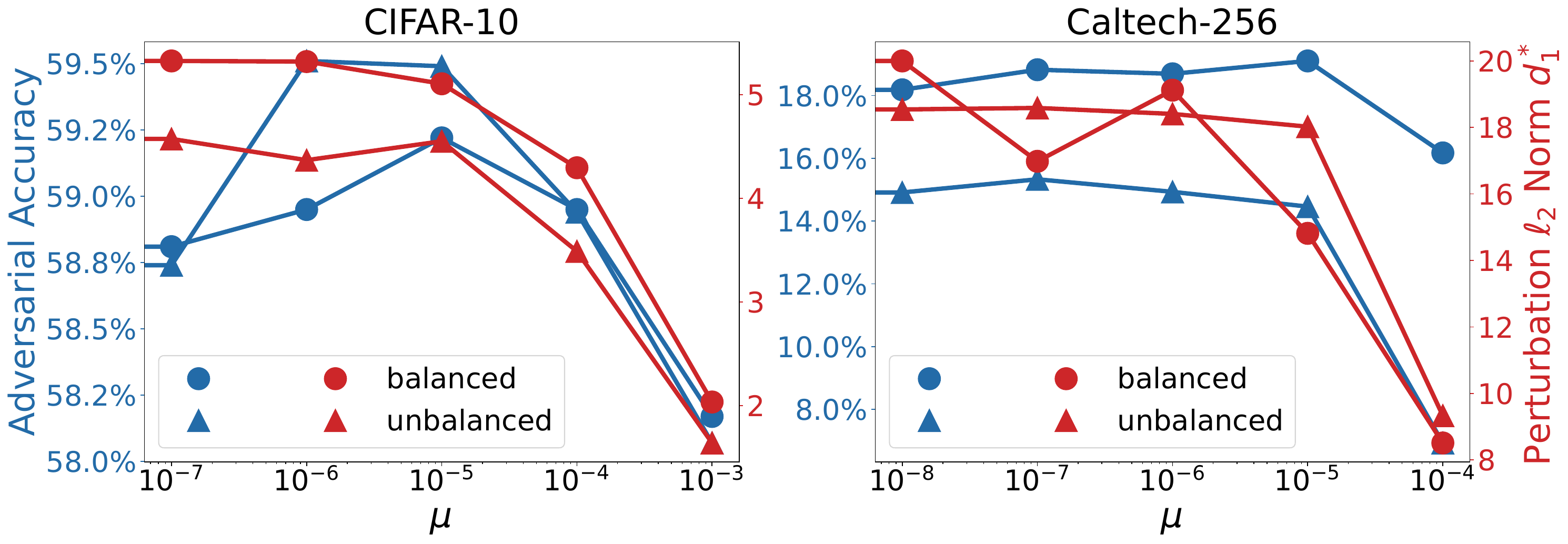}
\vspace{-2em}
\caption{Influence of the strong convexity hyperparameter $\mu$ on adversarial accuracy and perturbation magnitude. }
\label{fig:exp_mu}
\vspace{-1em}
\end{figure}

\paragraph{Local Trainer with Strong Convexity Regularization.} 
We discuss how the strong convexity regularization in the local trainer of each client influences the robustness of \design. \cref{fig:exp_mu} shows the adversarial accuracy (in blue color) and the $\ell_2$ magnitude of the perturbation $\Delta\vz_1$, with respect to different strong convexity hyperparameter $\mu$. We can see that increasing $\mu$ in the range of $[0,10^{-5}]$ slightly increases the adversarial accuracy and decreases the perturbation magnitude. The insignificant change is attributed to the local strong convexity of the fully connected layer used in the auxiliary model. Though the fully connected layer followed by the cross-entropy loss (or equivalently, multinomial logistic regression) only has convexity instead of strong convexity globally, it can still be locally strongly convex with most inputs \cite{bohning1992multinomial}. Thus the strong convexity hyperparameter $\mu$ does not make a significant difference when it is small. When further increasing $\mu$, the perturbation magnitude begins to drop significantly, which is aligned with the conclusion of \cref{lemma:robustness}. However, using an over-large regularization ($\mu\ge 10^{-4}$) can also distract the training and even lead to divergence.

\begin{figure}[t]
\centering
\includegraphics[width=\linewidth]{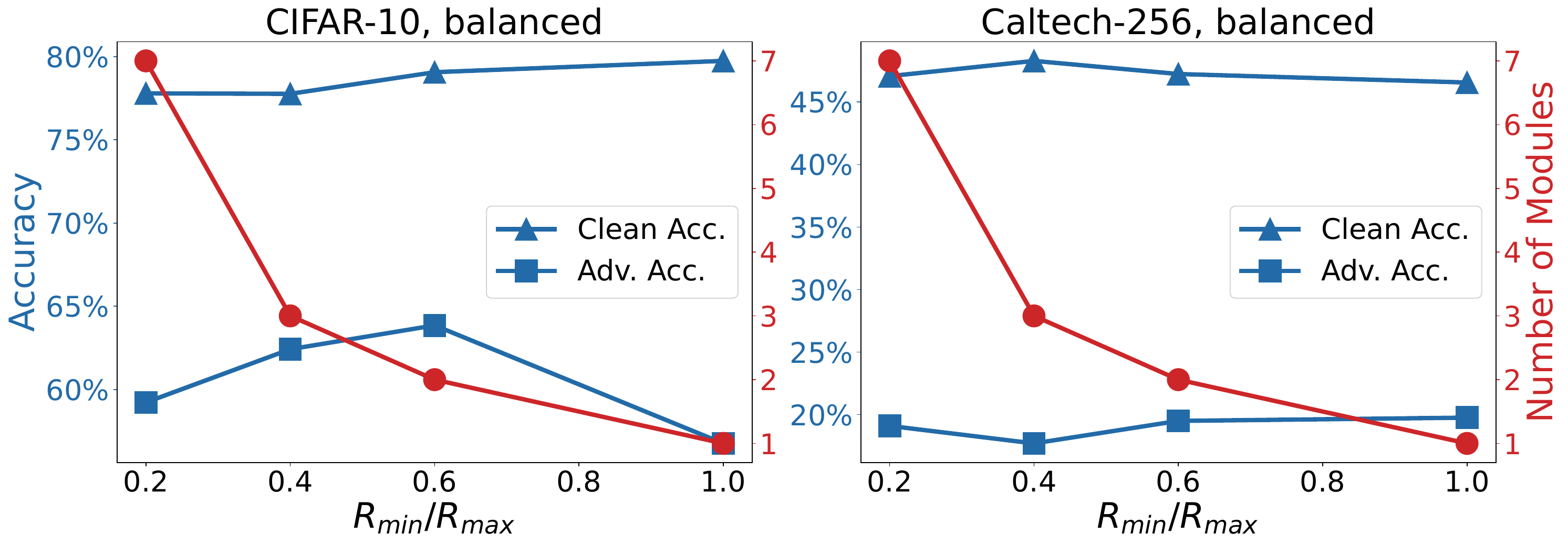}
\vspace{-2em}
\caption{The number of modules and the clean/adversarial accuracy with different $R_{\text{min}}$ (given in the ratio of $R_{\text{max}}$). }
\vspace{-1em}
\label{fig:exp_r}
\end{figure}

\begin{table}[t]
\centering
\caption{Performance with or without adaptive perturbation adjustment (APA) and differentiated module assignment (DMA).}
\label{table:ablation}
\resizebox{\linewidth}{!}{
\begin{tabular}{cc|cc|cc}
\hline
\multicolumn{2}{c|}{Dataset}      & \multicolumn{2}{c|}{CIFAR-10}                                                 & \multicolumn{2}{c}{Caltech-256}                                              \\
\multicolumn{2}{c|}{Sys. Hetero.} & \multicolumn{1}{c|}{balanced}              & \multicolumn{1}{c|}{unbalanced} & \multicolumn{1}{c|}{balanced}              & \multicolumn{1}{c}{unbalanced} \\ \hline
APA              & DMA        & \multicolumn{1}{c|}{Clean / Adv. Acc.} & \multicolumn{1}{c|}{Clean / Adv. Acc.} & \multicolumn{1}{c|}{Clean / Adv. Acc.} & \multicolumn{1}{c}{Clean / Adv. Acc.}    \\ \hline

\ding{51}              & \ding{51}       & \multicolumn{1}{c|}{77.79\% / 59.22\%}   & \multicolumn{1}{c|}{76.47\% / 59.51\%}      & \multicolumn{1}{c|}{45.04\% / 19.74\%}  & \multicolumn{1}{c}{43.39\% / 14.93\%}       \\

\ding{55}               & \ding{51}    & \multicolumn{1}{c|}{79.04\% / 56.98\%}    & \multicolumn{1}{c|}{77.02\% / 58.01\%}        & \multicolumn{1}{c|}{59.99\% / 10.80\%}   & \multicolumn{1}{c}{53.64\% / ~~8.06\%}     \\

\ding{51}              & \ding{55}     & \multicolumn{1}{c|}{71.66\% / 57.18\%}   & \multicolumn{1}{c|}{71.66\% / 57.18\%}  & \multicolumn{1}{c|}{14.67\% / ~~7.93\%}    & \multicolumn{1}{c}{14.67\% / ~~7.93\%}        \\

\ding{55}               & \ding{55}   & \multicolumn{1}{c|}{71.68\% / 57.34\%}   & \multicolumn{1}{c|}{71.68\% / 57.34\%}      & \multicolumn{1}{c|}{25.17\% / ~~4.38\%}    & \multicolumn{1}{c}{25.17\% / ~~4.38\%}  \\ \hline

\end{tabular}
}
\vspace{-0.5em}
\end{table}

\paragraph{Model Partitioner.} \cref{fig:exp_r} shows the number of modules and the corresponding performance of \design when partitioning the backbone model with different $R_{\text{min}}$. When the clients have more available memory, the number of modules decreases and finally degenerates to jFAT with only one module. However, the performance of \design does not show much difference with different numbers of modules, which also implies the effectiveness of our inconsistency-reduction designs in \design. When training with more than one module in \design, our adversarial cascade learning with strong convexity regularization guarantees the sufficient condition for robustness, thus leading to even higher adversarial accuracy than joint federated adversarial training in some cases.

\begin{figure}[t]
\centering
\includegraphics[width=\linewidth]{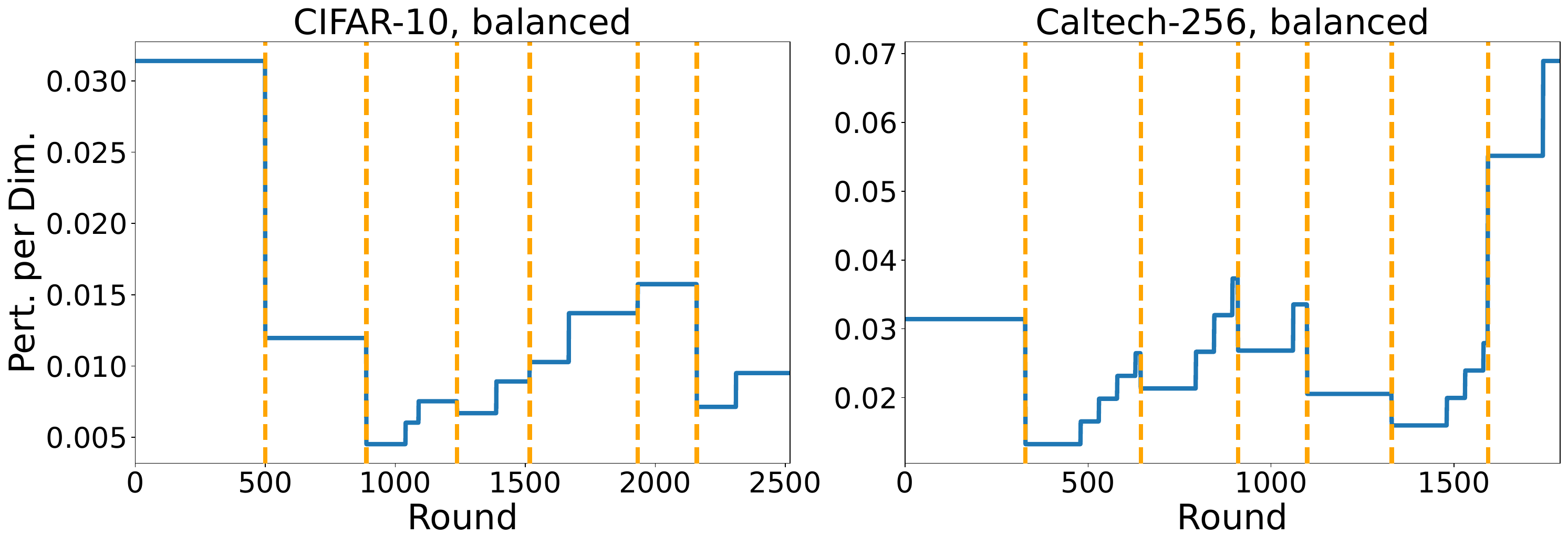}
\vspace{-2em}
\caption{Perturbation magnitude per dimension during the training of \design with adaptive perturbation adjustment in the balanced setting. The orange dash lines divide the training stages of each module $m\in\{1,2,\cdots,7\}$.}
\label{fig:exp_pert}
\vspace{-1em}
\end{figure}

\begin{table}[t]
\centering
\caption{Training time with or without DMA in \design.}
\label{table:dma_time}
\resizebox{\linewidth}{!}{
\begin{tabular}{c|cc|cc}
\hline
Dataset      & \multicolumn{2}{c|}{CIFAR-10}                              & \multicolumn{2}{c}{Caltech-256}                            \\
Sys. Hetero. & \multicolumn{1}{c|}{balanced}          & unbalanced        & \multicolumn{1}{c|}{balanced}          & unbalanced        \\ \hline
w/ DMA       & \multicolumn{1}{c|}{$9.2\times 10^4$s} & $1.8\times 10^5$s & \multicolumn{1}{c|}{$3.6\times 10^4$s} & $6.2\times 10^4$s \\
w/o DMA      & \multicolumn{1}{c|}{$9.1\times 10^4$s} & $1.9\times 10^5$s & \multicolumn{1}{c|}{$4.2\times 10^4$s} & $6.5\times 10^4$s \\ \hline
\end{tabular}
}
\vspace{-0.5em}
\end{table}

\paragraph{Training Coordinator.}
\cref{table:ablation} shows the performance of \design with/without Adaptive Perturbation Adjustment and Differentiated Module Assignment in the training coordinator. We can get the following two conclusions:
\begin{asparadesc}
\item[(a)] \textbf{Adaptive Perturbation Adjustment improves robustness and attains better utility-robustness trade-off.} When training without Adaptive Perturbation Adjustment (APA), \design achieves higher clean accuracy but lower adversarial accuracy. The robustness-utility ratio is lower than \design with APA and jFAT, especially when on Caltech-256. \cref{fig:exp_pert} shows that the perturbation magnitude starts from a relatively small value and increases gradually by APA when training each module. We find that initializing $\alpha_m$ with a small value (e.g., $0.3$) can stabilize the adversarial training, while APA can adjust the perturbation magnitude to achieve a better balance between the clean accuracy and the adversarial accuracy by comparing the accuracy ratio with the previous module.

\item[(b)] \textbf{Differentiated Module Assignment significantly improves the performance of \design.} We can see that both clean and adversarial accuracy drops when removing Differentiated Module Assignment (DMA) from \design, especially on Caltech-256 where the adversarial accuracy is not high enough to mitigate the objective inconsistency with the robustness-consistency relationship in \cref{lemma:consist}. DMA fully utilizes the resources of each client and assigns ``prophet'' clients more modules for training, such that the ``prophet'' clients can train more modules jointly with the ``future'' loss that is closer to the final loss of the whole backbone model. And \cref{table:dma_time} shows that DMA does not incur extra training latency with the FLOPs constraint in \cref{eq:flop_cons}, which avoids enlarging the synchronization time in each communication round of FL. 
\end{asparadesc}

\section{Conclusions and Future Works}\label{sec:con}
We propose \design, a memory-efficient federated adversarial training framework with robust and consistent cascade learning. We develop both the client-side local trainer and the server-side training coordinator to achieve high utility and strong robustness simultaneously. On the client side, We theoretically analyze the robustness condition and propose adversarial cascade learning with strong convexity regularization to guarantee the robustness of the backbone model. We further reveal that the robustness achieved in adversarial cascade learning also implies low objective inconsistency, leading to high utility at the same time. On the server side, we propose the memory-constrained model partitioner to automatically partition a given model to fit into the memory constraints among clients. We derive a training coordinator with Adaptive Perturbation Adjustment and Differentiated Module Assignment on the server to achieve the optimal utility-robustness trade-off and further reduce the objective inconsistency. Our empirical results show that \design consistently outperforms other memory-efficient federated learning methods. Compared with joint training, \design maintains comparable utility and robustness, with significant memory reduction and training speedup.

One future work would be extending \design to NLP tasks. Although the adversarial robustness of NLP tasks is less straightforward since it is difficult to generate adversarial examples with gradient-ascent methods (like PGD and AutoAttack) on the discrete texts, previous studies show that training with adversarially perturbed token embeddings can improve the generalization ability of the language model \cite{zhu2019freelb}. Therefore, it is meaningful to explore how \design can be applied with language models like Transformers \cite{vaswani2017attention} to achieve memory efficiency and generalization in NLP tasks.

Another potential direction is to combine \design with other memory-efficient training methods, like low-bit training \cite{zhong2022exploring} and LoRA \cite{hu2021lora}. Since \design partitions the backbone model with a layer or a block as the ``atom'', it is complementary to the parameter-level quantization and the layer-level low-rank approximation. Thus, \design can be applied with the other parameter-level or layer-level memory-efficient training methods to further reduce the memory requirement.

\section*{Acknowledgements}
We appreciate the constructive comments of the reviewers. 
This research is generously supported in part by Gift from Accenture, NSF CNS-2112562, CNS-2233808, CNS-2148253, and ARO W911NF-23-2-0224.

\bibliographystyle{mlsys2025}
\bibliography{references}

\begin{thebibliography}{44}
\providecommand{\natexlab}[1]{#1}
\providecommand{\url}[1]{\texttt{#1}}
\expandafter\ifx\csname urlstyle\endcsname\relax
  \providecommand{\doi}[1]{doi: #1}\else
  \providecommand{\doi}{doi: \begingroup \urlstyle{rm}\Url}\fi

\bibitem[Alam et~al.(2022)Alam, Liu, Yan, and Zhang]{alam2022fedrolex}
Alam, S., Liu, L., Yan, M., and Zhang, M.
\newblock Fedrolex: Model-heterogeneous federated learning with rolling sub-model extraction.
\newblock \emph{Advances in neural information processing systems}, 35:\penalty0 29677--29690, 2022.

\bibitem[Belilovsky et~al.(2020)Belilovsky, Eickenberg, and Oyallon]{belilovsky2020decoupled}
Belilovsky, E., Eickenberg, M., and Oyallon, E.
\newblock Decoupled greedy learning of cnns.
\newblock In \emph{International Conference on Machine Learning}, pp.\  736--745. PMLR, 2020.

\bibitem[B{\"o}hning(1992)]{bohning1992multinomial}
B{\"o}hning, D.
\newblock Multinomial logistic regression algorithm.
\newblock \emph{Annals of the institute of Statistical Mathematics}, 44\penalty0 (1):\penalty0 197--200, 1992.

\bibitem[Caldas et~al.(2018)Caldas, Kone{\v{c}}ny, McMahan, and Talwalkar]{caldas2018expanding}
Caldas, S., Kone{\v{c}}ny, J., McMahan, H.~B., and Talwalkar, A.
\newblock Expanding the reach of federated learning by reducing client resource requirements.
\newblock \emph{arXiv preprint arXiv:1812.07210}, 2018.

\bibitem[Cho et~al.(2022)Cho, Manoel, Joshi, Sim, and Dimitriadis]{cho2022heterogeneous}
Cho, Y.~J., Manoel, A., Joshi, G., Sim, R., and Dimitriadis, D.
\newblock Heterogeneous ensemble knowledge transfer for training large models in federated learning.
\newblock \emph{arXiv preprint arXiv:2204.12703}, 2022.

\bibitem[Croce \& Hein(2020)Croce and Hein]{croce2020reliable}
Croce, F. and Hein, M.
\newblock Reliable evaluation of adversarial robustness with an ensemble of diverse parameter-free attacks.
\newblock In \emph{International conference on machine learning}, pp.\  2206--2216. PMLR, 2020.

\bibitem[Diao et~al.(2020)Diao, Ding, and Tarokh]{diao2020heterofl}
Diao, E., Ding, J., and Tarokh, V.
\newblock Heterofl: Computation and communication efficient federated learning for heterogeneous clients.
\newblock \emph{arXiv preprint arXiv:2010.01264}, 2020.

\bibitem[Goodfellow et~al.(2014)Goodfellow, Shlens, and Szegedy]{goodfellow2014explaining}
Goodfellow, I.~J., Shlens, J., and Szegedy, C.
\newblock Explaining and harnessing adversarial examples.
\newblock \emph{arXiv preprint arXiv:1412.6572}, 2014.

\bibitem[Griffin et~al.(2007)Griffin, Holub, and Perona]{griffin2007caltech}
Griffin, G., Holub, A., and Perona, P.
\newblock Caltech-256 object category dataset.
\newblock 2007.

\bibitem[He et~al.(2016)He, Zhang, Ren, and Sun]{he2016deep}
He, K., Zhang, X., Ren, S., and Sun, J.
\newblock Deep residual learning for image recognition.
\newblock In \emph{Proceedings of the IEEE conference on computer vision and pattern recognition}, 2016.

\bibitem[Hettinger et~al.(2017)Hettinger, Christensen, Ehlert, Humpherys, Jarvis, and Wade]{hettinger2017forward}
Hettinger, C., Christensen, T., Ehlert, B., Humpherys, J., Jarvis, T., and Wade, S.
\newblock Forward thinking: Building and training neural networks one layer at a time.
\newblock \emph{arXiv preprint arXiv:1706.02480}, 2017.

\bibitem[Hong et~al.(2023)Hong, Wang, Wang, and Zhou]{hong2023federated}
Hong, J., Wang, H., Wang, Z., and Zhou, J.
\newblock Federated robustness propagation: sharing adversarial robustness in heterogeneous federated learning.
\newblock In \emph{Proceedings of the AAAI Conference on Artificial Intelligence}, volume~37, pp.\  7893--7901, 2023.

\bibitem[Hu et~al.(2021)Hu, Shen, Wallis, Allen-Zhu, Li, Wang, Wang, and Chen]{hu2021lora}
Hu, E.~J., Shen, Y., Wallis, P., Allen-Zhu, Z., Li, Y., Wang, S., Wang, L., and Chen, W.
\newblock Lora: Low-rank adaptation of large language models.
\newblock \emph{arXiv preprint arXiv:2106.09685}, 2021.

\bibitem[Kairouz et~al.(2019)Kairouz, McMahan, Avent, Bellet, Bennis, Bhagoji, Bonawitz, Charles, Cormode, Cummings, et~al.]{kairouz2019advances}
Kairouz, P., McMahan, H.~B., Avent, B., Bellet, A., Bennis, M., Bhagoji, A.~N., Bonawitz, K., Charles, Z., Cormode, G., Cummings, R., et~al.
\newblock Advances and open problems in federated learning.
\newblock \emph{arXiv preprint arXiv:1912.04977}, 2019.

\bibitem[Karimireddy et~al.(2019)Karimireddy, Kale, Mohri, Reddi, Stich, and Suresh]{karimireddy2019scaffold}
Karimireddy, S.~P., Kale, S., Mohri, M., Reddi, S.~J., Stich, S.~U., and Suresh, A.~T.
\newblock Scaffold: Stochastic controlled averaging for on-device federated learning.
\newblock \emph{arXiv preprint arXiv:1910.06378}, 2019.

\bibitem[Kone{\v{c}}n{\`y} et~al.(2015)Kone{\v{c}}n{\`y}, McMahan, and Ramage]{konevcny2015federated}
Kone{\v{c}}n{\`y}, J., McMahan, B., and Ramage, D.
\newblock Federated optimization: Distributed optimization beyond the datacenter.
\newblock \emph{arXiv preprint arXiv:1511.03575}, 2015.

\bibitem[Kone{\v{c}}n{\`y} et~al.(2016)Kone{\v{c}}n{\`y}, McMahan, Yu, Richt{\'a}rik, Suresh, and Bacon]{konevcny2016federated}
Kone{\v{c}}n{\`y}, J., McMahan, H.~B., Yu, F.~X., Richt{\'a}rik, P., Suresh, A.~T., and Bacon, D.
\newblock Federated learning: Strategies for improving communication efficiency.
\newblock \emph{arXiv preprint arXiv:1610.05492}, 2016.

\bibitem[Krizhevsky et~al.(2009)Krizhevsky, Hinton, et~al.]{krizhevsky2009learning}
Krizhevsky, A., Hinton, G., et~al.
\newblock Learning multiple layers of features from tiny images.
\newblock 2009.

\bibitem[Li et~al.(2018)Li, Sahu, Zaheer, Sanjabi, Talwalkar, and Smith]{li2018federated}
Li, T., Sahu, A.~K., Zaheer, M., Sanjabi, M., Talwalkar, A., and Smith, V.
\newblock Federated optimization in heterogeneous networks.
\newblock \emph{arXiv preprint arXiv:1812.06127}, 2018.

\bibitem[Li et~al.(2020)Li, Sahu, Talwalkar, and Smith]{li2020federated}
Li, T., Sahu, A.~K., Talwalkar, A., and Smith, V.
\newblock Federated learning: Challenges, methods, and future directions.
\newblock \emph{IEEE Signal Processing Magazine}, 37\penalty0 (3):\penalty0 50--60, 2020.

\bibitem[Lin et~al.(2020)Lin, Kong, Stich, and Jaggi]{lin2020ensemble}
Lin, T., Kong, L., Stich, S.~U., and Jaggi, M.
\newblock Ensemble distillation for robust model fusion in federated learning.
\newblock \emph{Advances in Neural Information Processing Systems}, 33:\penalty0 2351--2363, 2020.

\bibitem[Madry et~al.(2017)Madry, Makelov, Schmidt, Tsipras, and Vladu]{madry2017towards}
Madry, A., Makelov, A., Schmidt, L., Tsipras, D., and Vladu, A.
\newblock Towards deep learning models resistant to adversarial attacks.
\newblock \emph{arXiv preprint arXiv:1706.06083}, 2017.

\bibitem[Marquez et~al.(2018)Marquez, Hare, and Niranjan]{marquez2018deep}
Marquez, E.~S., Hare, J.~S., and Niranjan, M.
\newblock Deep cascade learning.
\newblock \emph{IEEE transactions on neural networks and learning systems}, 29\penalty0 (11):\penalty0 5475--5485, 2018.

\bibitem[McMahan et~al.(2017)McMahan, Moore, Ramage, Hampson, and y~Arcas]{mcmahan2017communication}
McMahan, B., Moore, E., Ramage, D., Hampson, S., and y~Arcas, B.~A.
\newblock Communication-efficient learning of deep networks from decentralized data.
\newblock In \emph{Artificial Intelligence and Statistics}, pp.\  1273--1282. PMLR, 2017.

\bibitem[Moosavi-Dezfooli et~al.(2019)Moosavi-Dezfooli, Fawzi, Uesato, and Frossard]{moosavi2019robustness}
Moosavi-Dezfooli, S.-M., Fawzi, A., Uesato, J., and Frossard, P.
\newblock Robustness via curvature regularization, and vice versa.
\newblock In \emph{Proceedings of the IEEE/CVF Conference on Computer Vision and Pattern Recognition}, pp.\  9078--9086, 2019.

\bibitem[Rajbhandari et~al.(2020)Rajbhandari, Rasley, Ruwase, and He]{rajbhandari2020zero}
Rajbhandari, S., Rasley, J., Ruwase, O., and He, Y.
\newblock Zero: Memory optimizations toward training trillion parameter models.
\newblock In \emph{SC20: International Conference for High Performance Computing, Networking, Storage and Analysis}, pp.\  1--16. IEEE, 2020.

\bibitem[Shah et~al.(2021)Shah, Dube, Chakraborty, and Verma]{shah2021adversarial}
Shah, D., Dube, P., Chakraborty, S., and Verma, A.
\newblock Adversarial training in communication constrained federated learning.
\newblock \emph{arXiv preprint arXiv:2103.01319}, 2021.

\bibitem[Simonyan \& Zisserman(2014)Simonyan and Zisserman]{simonyan2014very}
Simonyan, K. and Zisserman, A.
\newblock Very deep convolutional networks for large-scale image recognition.
\newblock \emph{arXiv preprint arXiv:1409.1556}, 2014.

\bibitem[Sun et~al.(2022)Sun, Li, Duan, Alam, Deng, Guo, Wang, Gorlatova, Zhang, Li, et~al.]{sun2022fedsea}
Sun, J., Li, A., Duan, L., Alam, S., Deng, X., Guo, X., Wang, H., Gorlatova, M., Zhang, M., Li, H., et~al.
\newblock Fedsea: A semi-asynchronous federated learning framework for extremely heterogeneous devices.
\newblock In \emph{Proceedings of the 20th ACM Conference on Embedded Networked Sensor Systems}, pp.\  106--119, 2022.

\bibitem[Tang et~al.(2022)Tang, Ning, Wang, Sun, Wang, Li, and Chen]{tang2022fedcor}
Tang, M., Ning, X., Wang, Y., Sun, J., Wang, Y., Li, H., and Chen, Y.
\newblock Fedcor: Correlation-based active client selection strategy for heterogeneous federated learning.
\newblock In \emph{Proceedings of the IEEE/CVF Conference on Computer Vision and Pattern Recognition}, pp.\  10102--10111, 2022.

\bibitem[Tian et~al.(2022)Tian, Li, Shi, Wang, and Xu]{tian2022harmony}
Tian, C., Li, L., Shi, Z., Wang, J., and Xu, C.
\newblock Harmony: Heterogeneity-aware hierarchical management for federated learning system.
\newblock In \emph{2022 55th IEEE/ACM International Symposium on Microarchitecture (MICRO)}, pp.\  631--645. IEEE, 2022.

\bibitem[Vaswani(2017)]{vaswani2017attention}
Vaswani, A.
\newblock Attention is all you need.
\newblock \emph{Advances in Neural Information Processing Systems}, 2017.

\bibitem[Wang et~al.(2023)Wang, Qin, Jacobs, Holmes, Rajbhandari, Ruwase, Yan, Yang, and He]{wang2023zero++}
Wang, G., Qin, H., Jacobs, S.~A., Holmes, C., Rajbhandari, S., Ruwase, O., Yan, F., Yang, L., and He, Y.
\newblock Zero++: Extremely efficient collective communication for giant model training.
\newblock \emph{arXiv preprint arXiv:2306.10209}, 2023.

\bibitem[Wang et~al.(2020)Wang, Liu, Liang, Joshi, and Poor]{wang2020tackling}
Wang, J., Liu, Q., Liang, H., Joshi, G., and Poor, H.~V.
\newblock Tackling the objective inconsistency problem in heterogeneous federated optimization.
\newblock \emph{arXiv preprint arXiv:2007.07481}, 2020.

\bibitem[Wang et~al.(2021{\natexlab{a}})Wang, Ma, Bailey, Yi, Zhou, and Gu]{wang2021convergence}
Wang, Y., Ma, X., Bailey, J., Yi, J., Zhou, B., and Gu, Q.
\newblock On the convergence and robustness of adversarial training.
\newblock \emph{arXiv preprint arXiv:2112.08304}, 2021{\natexlab{a}}.

\bibitem[Wang et~al.(2021{\natexlab{b}})Wang, Ni, Song, Yang, and Huang]{wang2021revisiting}
Wang, Y., Ni, Z., Song, S., Yang, L., and Huang, G.
\newblock Revisiting locally supervised learning: An alternative to end-to-end training.
\newblock \emph{arXiv preprint arXiv:2101.10832}, 2021{\natexlab{b}}.

\bibitem[Wen et~al.(2022)Wen, Jeon, and Huang]{wen2022federated}
Wen, D., Jeon, K.-J., and Huang, K.
\newblock Federated dropout—a simple approach for enabling federated learning on resource constrained devices.
\newblock \emph{IEEE wireless communications letters}, 11\penalty0 (5):\penalty0 923--927, 2022.

\bibitem[Wong et~al.(2020)Wong, Rice, and Kolter]{wong2020fast}
Wong, E., Rice, L., and Kolter, J.~Z.
\newblock Fast is better than free: Revisiting adversarial training.
\newblock \emph{arXiv preprint arXiv:2001.03994}, 2020.

\bibitem[Yang et~al.(2020)Yang, Zhang, Dong, Inkawhich, Gardner, Touchet, Wilkes, Berry, and Li]{yang2020dverge}
Yang, H., Zhang, J., Dong, H., Inkawhich, N., Gardner, A., Touchet, A., Wilkes, W., Berry, H., and Li, H.
\newblock Dverge: diversifying vulnerabilities for enhanced robust generation of ensembles.
\newblock \emph{Advances in Neural Information Processing Systems}, 33:\penalty0 5505--5515, 2020.

\bibitem[Zhang et~al.(2023)Zhang, Li, Tang, Sun, Chen, Zhang, Chen, Chen, and Li]{zhang2023fed}
Zhang, J., Li, A., Tang, M., Sun, J., Chen, X., Zhang, F., Chen, C., Chen, Y., and Li, H.
\newblock Fed-cbs: A heterogeneity-aware client sampling mechanism for federated learning via class-imbalance reduction.
\newblock In \emph{International Conference on Machine Learning}, pp.\  41354--41381. PMLR, 2023.

\bibitem[Zhang \& Zhu(2019)Zhang and Zhu]{zhang2019interpreting}
Zhang, T. and Zhu, Z.
\newblock Interpreting adversarially trained convolutional neural networks.
\newblock In \emph{International conference on machine learning}, pp.\  7502--7511. PMLR, 2019.

\bibitem[Zhong et~al.(2022)Zhong, Ning, Dai, Zhu, Zhao, Zeng, Wang, and Yang]{zhong2022exploring}
Zhong, K., Ning, X., Dai, G., Zhu, Z., Zhao, T., Zeng, S., Wang, Y., and Yang, H.
\newblock Exploring the potential of low-bit training of convolutional neural networks.
\newblock \emph{IEEE Transactions on Computer-Aided Design of Integrated Circuits and Systems}, 41\penalty0 (12):\penalty0 5421--5434, 2022.

\bibitem[Zhu et~al.(2019)Zhu, Cheng, Gan, Sun, Goldstein, and Liu]{zhu2019freelb}
Zhu, C., Cheng, Y., Gan, Z., Sun, S., Goldstein, T., and Liu, J.
\newblock Freelb: Enhanced adversarial training for natural language understanding.
\newblock \emph{arXiv preprint arXiv:1909.11764}, 2019.

\bibitem[Zizzo et~al.(2020)Zizzo, Rawat, Sinn, and Buesser]{zizzo2020fat}
Zizzo, G., Rawat, A., Sinn, M., and Buesser, B.
\newblock Fat: Federated adversarial training.
\newblock \emph{arXiv preprint arXiv:2012.01791}, 2020.

\end{thebibliography}

\clearpage
\appendix
\section{Proofs}
\setcounter{lemma}{0}
\subsection{Proof of Lemma 1}\label{proof:lemma_robustness}
\begin{lemma}
    If $l_m$ is $\mu_m$-strongly convex on $\vz_m$ and module $m$ is $(\epsilon_{m-1},c_m)$-robust in $l_m$, we have
    \begin{equation*}
        \begin{aligned}
        &\max_{\Vert\boldsymbol{\delta}_{m-1}\Vert\le \epsilon_{m-1}}\Vert \Delta \vz_m\Vert_2\\ 
        \le &\frac{\Vert \nabla_{\vz_m} l_m(\vz_m,y)\Vert_2}{\mu_m}+\sqrt{\frac{2c_m}{\mu_m}+\frac{\Vert \nabla_{\vz_m} l_m(\vz_m,y)\Vert_2^2}{\mu_m^2}}.
        \end{aligned}
    \end{equation*}
    \begin{proof}
        With the strong convexity and the $(\epsilon_{m-1},c_m)$-robustness of $l_m$, we have
        \begin{align*}
            &(\nabla_{\vz_m}l_m(\vz_m,y))^T\Delta \vz_m+\frac{\mu_m}{2}\Vert \Delta \vz_m\Vert_2^2\\
            &\le l_m(\vz_m+\Delta \vz_m,y)- l_m(\vz_m,y)\le c_m\\
            \Rightarrow &\left\Vert\Delta \vz_m+\frac{\nabla_{\vz_m}l_m(\vz_m,y)}{\mu_m} \right\Vert_2\\
            &\qquad\qquad\quad\le \sqrt{\frac{2c_m}{\mu_m}+\frac{\Vert \nabla_{\vz_m} l_m(\vz_m,y)\Vert_2^2}{\mu_m^2}}\\
            \Rightarrow &\Vert \Delta \vz_m \Vert_2 \le \frac{\Vert\nabla_{\vz_m} l_m(\vz_m,y)\Vert_2}{\mu_m}\\
            &\qquad\qquad\quad+\sqrt{\frac{2c_m}{\mu_m}+\frac{\Vert \nabla_{\vz_m} l_m(\vz_m,y)\Vert_2^2}{\mu_m^2}}.
        \end{align*}
    \end{proof}
\end{lemma}

\subsection{Proof of Lemma 2}\label{proof:lemma_consist}
\begin{lemma}
    If the early exit loss $l_m$ has $\beta_m$-smoothness and $(\epsilon_m,c_m)$-robustness on $\vz_m$, the joint loss $l$ has $\beta_m'$-smoothness and $(\epsilon_m,c_M)$-robustness on $\vz_m$, we have
    \begin{equation*}
    \begin{aligned}
        &\Vert\nabla_{\vw_m} l-\nabla_{\vw_m} l_m\Vert_2\\
        \le&\left\Vert \frac{\partial \vz_m}{\partial \vw_m}\right\Vert_2\sqrt{2(c_m+c_M)(\beta_m+\beta_m')}.
    \end{aligned}
    \end{equation*}
    \begin{proof}
        We define $h_m(\vz_m) = l(\vz_m)-l_m(\vz_m)$, which has $(\beta_m+\beta_m')$-smoothness and $(\epsilon_m,c_m+c_M)$-robustness on $\vz_m$. Thus $\forall\boldsymbol{\delta}_m$ with $\Vert \boldsymbol{\delta}_m\Vert\le \epsilon_m$, we have
        \begin{align*}
            &\left(\nabla_{\vz_m} h_m(\vz_m)\right)^T\boldsymbol{\delta}_m-\frac{\beta_m+\beta_m'}{2}\Vert \boldsymbol{\delta}_m\Vert_2^2\\
            &\le h_m(\vz_m+\boldsymbol{\delta}_m)-h_m(\vz_m)\le  c_m+ c_M.
        \end{align*}
        We take the maximum of the left hand side with $\boldsymbol{\delta}^*_m=\frac{\nabla_{\vz_m}h_m(\vz_m)}{\beta_m+\beta_m'}$, and we can get
        \begin{align*}
            &\frac{\left\Vert\nabla_{\vz_m}h_m(\vz_m)\right\Vert_2^2}{2(\beta_m+\beta_m')}\le c_m+c_M\\
            \Rightarrow& \left\Vert\nabla_{\vz_m}h_m(\vz_m)\right\Vert_2\le \sqrt{2(c_m+c_M)(\beta_m+\beta_m')}.
        \end{align*}
        With the chain rule and $\Vert\nabla_{\vw_m} l-\nabla_{\vw_m} l_m\Vert_2\le \left\Vert \frac{\partial \vz_m}{\partial \vw_m}\right\Vert_2\Vert \nabla_{\vz_m} l-\nabla_{\vz_m} l_m\Vert_2$, we prove the lemma.
    \end{proof}
\end{lemma}

\section{Experiment Details}\label{apx:exp}
\subsection{Device Details}\label{apx:dev}
Considering the different memory and performance requirements for training on CIFAR-10 (small images) and Caltech-256 (large images), we collect two device pools for CIFAR-10 (\cref{table:cifar_dev}) and Caltech-256 (\cref{table:cal_dev}) respectively. Meanwhile, we multiply \textit{degrading factors} to the peak memory and performance to simulate the real-time available memory and performance of each client with different co-running runtime applications, such as 4k-video playing and object detection \cite{tian2022harmony}. Specifically, the degrading factor for memory is uniformly sampled from $[0,0.2]$, and the factor for performance is uniformly sampled from $[0,1.0]$.

\begin{table}[t]
\centering
\caption{Performance, memory, and storage I/O Bandwidth of the devices for training on CIFAR-10.}
\label{table:cifar_dev}
\resizebox{\linewidth}{!}{
\begin{tabular}{c|ccc}
\hline
Device          & Performance                & Memory        & I/O Bandwidth            \\ \hline
GTX 1650m       & 3.1 TFLOPS                 & 4 GB          & 16 GB/s                  \\
TX2             & 1.3 TFLOPS                 & 4 GB          & 1.5 GB/s                 \\
KCU1500         & 0.2 TFLOPS                 & 2 GB          & 2 GB/s                   \\
VC709           & 0.1 TFLOPS                 & 2 GB          & 1.5 GB/s                 \\
Radeon HD 6870  & 2.7 TFLOPS                 & 1 GB          & 16 GB/s                  \\
Quadro M2200    & 2.1 TFLOPS                 & 4 GB          & 1.5 GB/s                 \\
A12 GPU         & 0.5 TFLOPS                 & 4 GB          & 1.5 GB/s                 \\
Geforce 750     & 1.1 TFLOPS                 & 1 GB          & 16 GB/s                  \\
Grid K240q      & 2.3 TFLOPS                 & 1 GB          & 16 GB/s                  \\
Radeon RX 6300m & 3.7 TFLOPS                 & 2 GB          & 16 GB/s                  \\ \hline
\end{tabular}
}
\vspace{-1.5em}
\end{table}

\begin{table}[t]
\centering
\caption{Performance, memory, and storage I/O Bandwidth of the devices for training on Caltech-256.}
\label{table:cal_dev}
\resizebox{\linewidth}{!}{
\begin{tabular}{c|ccc}
\hline
Device         & Performance          & Memory      & I/O Bandwidth     \\ \hline
Radeon RX 7600 & 21.8 TFLOPS          & 8 GB        & 16 GB/s           \\
Radeon RX 6800 & 16.2 TFLOPS          & 16 GB       & 16 GB/s           \\
Arc A770       & 19.7 TFLOPS          & 16 GB       & 16 GB/s           \\
Quadro P5000   & 5.3 TFLOPS           & 16 GB       & 1.5 GB/s          \\
RTX 3080m      & 19.0 TFLOPS          & 8 GB        & 16 GB/s           \\
RTX 4090m      & 33.0 TFLOPS          & 16 GB       & 16 GB/s           \\
A17 GPU        & 2.1 TFLOPS           & 8 GB        & 1.5 GB/s          \\
GTX 1650m      & 3.1 TFLOPS           & 4 GB        & 16 GB/s           \\
TX2            & 1.3 TFLOPS           & 4 GB        & 1.5 GB/s          \\
P104 101       & 8.6 TFLOPS           & 4 GB        & 16 GB/s           \\ \hline
\end{tabular}
}
\vspace{-1.5em}
\end{table}

\subsection{Baselines}\label{apx:base}
We compare \design with joint federated adversarial learning (jFAT) \cite{zizzo2020fat}, knowledge-distillation federated adversarial training (FedDF-AT \cite{lin2020ensemble}, FedET-AT \cite{cho2022heterogeneous}), partial-training federated adversarial training (HeteroFL-AT \cite{diao2020heterofl}, FedDrop-AT \cite{wen2022federated}, FedRolex-AT \cite{alam2022fedrolex}), and Federated Robustness Propagation (FedRBN) \cite{hong2023federated}.
\begin{asparadesc}
\item[(1)] jFAT trains the whole model end-to-end, with memory swapping if a client does not have sufficient memory. 

\item[(2)] In knowledge-distillation FL, each client selects the largest model that can be trained with the available memory from a group of models (\{CNN3, VGG11, VGG13, VGG16\} in CIFAR-10, \{CNN4, ResNet10, ResNet18, ResNet34\} in Caltech-256). The heterogeneous locally trained models are aggregated into the large global model by knowledge distillation with a small public dataset. 

\item[(3)] In partial-training FL, each client trains a sub-model of the whole model by dropping out a certain percentage of neurons or filters in each layer. The percentage is set as $1-R_k^{(t)}/R_{\text{max}}$ where $R_{\text{max}}$ is the memory requirement for training the whole model. 

\item[(4)] FedRBN allows clients with insufficient memory to conduct standard training only. The robustness is transferred from the batch normalization statistics of the memory-sufficient clients who conduct adversarial training to those who conduct standard training.
\end{asparadesc}

\subsection{Model Partition in \design}\label{apx:part}
According to \cref{alg:mp} and the minimal reserved memory in each setting, the VGG16 and ResNet34 are both partitioned into 7 modules as shown in \cref{table:vgg_part} and \cref{table:res_part}.

\begin{table}[t]
\centering
\caption{The model partition of VGG16 with $R_\text{min}=60$ MB. We show the memory requirement for training with SGD and the FLOPs of one forward propagation.}
\label{table:vgg_part}
\begin{tabular}{c|ccc}
\hline
Module             & Layer    & Mem. Req.             & FLOPs                 \\ \hline
\multirow{2}{*}{1} & Conv 1   & \multirow{2}{*}{55.8 MB} & \multirow{2}{*}{2.6 G} \\
                   & Conv 2   &                       &                       \\ \hline
\multirow{3}{*}{2} & Conv 3   & \multirow{3}{*}{46.1 MB} & \multirow{3}{*}{4.9 G} \\
                   & Conv 4   &                       &                       \\
                   & Conv 5   &                       &                       \\ \hline
\multirow{3}{*}{3} & Conv 6   & \multirow{3}{*}{50.4 MB} & \multirow{3}{*}{6.0 G} \\
                   & Conv 7   &                       &                       \\
                   & Conv 8   &                       &                       \\ \hline
4                  & Conv 9   & 34.7 MB                 & 2.4 G                  \\ \hline
5                  & Conv 10  & 33.1 MB                 & 2.4 G                  \\ \hline
\multirow{2}{*}{6} & Conv 11  & \multirow{2}{*}{59.3 MB} & \multirow{2}{*}{1.2 G} \\
                   & Conv 12  &                       &                       \\ \hline
\multirow{4}{*}{7} & Conv 13  & \multirow{4}{*}{36.1 MB} & \multirow{4}{*}{0.6 G} \\
                   & Linear 1 &                       &                       \\
                   & Linear 2 &                       &                       \\
                   & Linear 3 &                       &                       \\ \hline
\end{tabular}
\vspace{-0.8em}
\end{table}

\begin{table}[t]
\centering
\caption{The model partition of ResNet34 with $R_\text{min}=224$ MB. We show the memory requirement for training with SGD and the FLOPs of one forward propagation.}
\label{table:res_part}
\begin{tabular}{c|ccc}
\hline
Module             & Layer/Block   & Mem. Req.          & FLOPs                  \\ \hline
1                  & Conv          & 148.6 MB                 & 3.9 G                   \\ \hline
2                  & BasicBlock 1  & 130.2 MB                 & 7.5 G                   \\ \hline
3                  & BasicBlock 2  & 130.2 MB                 & 7.5 G                   \\ \hline
\multirow{2}{*}{4} & BasicBlock 3  & \multirow{2}{*}{197.9 MB} & \multirow{2}{*}{13.3 G} \\
                   & BasicBlock 4  &                        &                        \\ \hline
\multirow{4}{*}{5} & BasicBlock 5  & \multirow{4}{*}{221.6 MB} & \multirow{4}{*}{28.1 G} \\
                   & BasicBlock 6  &                        &                        \\
                   & BasicBlock 7  &                        &                        \\
                   & BasicBlock 8  &                        &                        \\ \hline
\multirow{5}{*}{6} & BasicBlock 9  & \multirow{5}{*}{206.5 MB} & \multirow{5}{*}{37.1 G} \\
                   & BasicBlock 10 &                        &                        \\
                   & BasicBlock 11 &                        &                        \\
                   & BasicBlock 12 &                        &                        \\
                   & BasicBlock 13 &                        &                        \\ \hline
\multirow{4}{*}{7} & BasicBlock 14 & \multirow{4}{*}{204.0 MB} & \multirow{4}{*}{20.6 G} \\
                   & BasicBlock 15 &                        &                        \\
                   & BasicBlock 16 &                        &                        \\
                   & Linear        &                        &                        \\ \hline
\end{tabular}
\vspace{-0.8em}
\end{table}

\subsection{Training Hyperparameters}
\paragraph{Common Hyperparameters} 
We conduct FL with $N=100$ clients, and we randomly select $C=10$ clients to participate in training at each communication round. To guarantee that each algorithm in \cref{table:main} and \cref{fig:exp_time} can converge, the total numbers of communication rounds are set to 500 for jFAT and 1000 for other baselines. In each communication round, each selected client conducts $E=30$ iterations of local SGD. The batch size is set to $B=64$ on CIFAR-10 and $B=32$ on Caltech-256, and the learning rates are $\eta_0=0.005$ and $0.001$ for VGG16 and ResNet34 respectively. We apply a learning rate decay factor $\gamma=0.994$ such that $\eta_t = \gamma^t\eta_0$ at communication round $t$. The momentum is set to be $0.9$, and the weight decay is set to be $10^{-4}$ in all our settings. 

\paragraph{Hyperparameters for Knowledge-distillation FL} We partition around 10\% of each dataset as the public dataset for knowledge distillation, namely, $5000$ samples in CIFAR-10 and $2500$ samples in Caltech-256. Following \citet{cho2022heterogeneous}, we set the iterations of distillation to be $128$, with the same learning rate and batch size in the common hyperparameters.

\paragraph{Hyperparameters for \design} We use $\mu=10^{-5}$ in \cref{table:main}, which is shown to be the optimal in \cref{fig:exp_mu}. We set $\gamma=0.05$ and $\Delta \alpha=0.1$ in all our experiments. We set the maximal number of communication rounds for each module to be $500$, while we allow \design to end the training of the current module early when the accuracy is not improved in the last $50$ rounds.

\end{document}